\documentclass[10pt,twocolumn]{IEEEtran}
\usepackage{graphicx}
\usepackage{epsfig}
\usepackage{amsmath}
\usepackage{bbm}
\usepackage{amssymb}
\usepackage{setspace}
\usepackage{wrapfig}
\usepackage{times,color}
\usepackage{caption}
\usepackage{subcaption}
\usepackage{algorithm}
\usepackage{algpseudocode}
\usepackage{balance}
\usepackage{epstopdf}
\usepackage{cite,footnote,xspace,syntonly,bm}

\usepackage{amsfonts}

\input{mysymbol.sty}
\allowdisplaybreaks[4]

\algnewcommand{\Inputs}[1]{%
  \State \textbf{Inputs:}
  \Statex \hspace*{\algorithmicindent}\parbox[t]{.8\linewidth}{\raggedright #1}
}
\algnewcommand{\Initialize}[1]{%
  \State \textbf{Initialize:}
  \Statex \hspace*{\algorithmicindent}\parbox[t]{.8\linewidth}{\raggedright #1}
}

\def \bbeta {{\boldsymbol \beta}}

\def \bbzeta {{\boldsymbol \zeta}}

\def \vec {\text{vec}}

\def \vec {\text{vec}}

 \long\def\symbolfootnote[#1]#2{\begingroup
 	\def\thefootnote{\fnsymbol{footnote}}
 	\footnote[#1]{#2}\endgroup} \psfull
 	 
\begin{document}

\title{\huge Tensor Decompositions for Identifying Directed Graph Topologies and Tracking Dynamic Networks}
\author{{{ Yanning Shen, \textit{Student Member}, \textit{IEEE}, Brian Baingana, \textit{Member}, \textit{IEEE},\\
 and Georgios~B.~Giannakis, \textit{Fellow, IEEE}}}\\
}

\markboth{IEEE TRANSACTIONS ON SIGNAL PROCESSING OCTOBER 26, 2016 \; (SUBMITTED)}{}

\maketitle  
\symbolfootnote[0]{$\dag$ Work in this paper was supported by grants NSF 1500713 and NIH 1R01GM104975-01.}

\symbolfootnote[0]{$\ast$ Y. Shen, B. Baingana and G. B. Giannakis are with the Dept.
	of ECE and the Digital Technology Center, University of
	Minnesota, 117 Pleasant Str. SE, Minneapolis, MN 55455. Tel: (612)625-4287; Emails:
	\texttt{\{shenx513,baing011,georgios\}@umn.edu }}



\vspace{-8mm}
\begin{abstract}	
Directed networks are pervasive both in nature and engineered systems, often underlying the complex behavior observed in biological systems, microblogs and social interactions over the web, as well as global financial markets. Since their structures are often unobservable, in order to facilitate network analytics, one generally resorts to approaches capitalizing on measurable nodal processes to infer the unknown topology. Structural equation models (SEMs) are capable of incorporating exogenous inputs to resolve inherent directional ambiguities. However, conventional SEMs assume full knowledge of exogenous inputs, which may not be readily available in some practical settings. The present paper advocates a novel SEM-based topology inference approach that entails factorization of a three-way tensor, constructed from the observed nodal data, using the well-known parallel factor (PARAFAC) decomposition. It turns out that second-order piecewise stationary statistics of exogenous variables suffice to identify the hidden topology. Capitalizing on the uniqueness properties inherent to high-order tensor factorizations, it is shown that topology identification is possible under reasonably mild conditions. In addition, to facilitate real-time operation and inference of time-varying networks, an adaptive (PARAFAC) tensor decomposition scheme which tracks the topology-revealing tensor factors is developed. Extensive tests on simulated and real stock quote data demonstrate the merits of the novel tensor-based approach. 
\end{abstract}



\section{Introduction}
\label{sec:intro}

The study of networks and network phenomena has recently emerged as a major catalyst for collectively understanding the behavior of complex systems \cite{kolaczyk2009statistical,easley2010networks,rogers2010diffusion}. Such systems are ubiquitous, and commonly arise in both natural and man-made settings. For example, online interactions over the web are commonly facilitated through social networks such as Facebook and Twitter, while sophisticated brain functions are the result of vast interactions within complex neuronal networks; see e.g.,~\cite{rubinov2010complex} and references therein. Other networks naturally emerge in settings as diverse as financial markets, genomics and proteomics, power grids, and transportation systems, to name just a few. 

While some of these networks are directly observable, due to e.g., presence of physical or engineered links between nodes, most complex networks have hidden topologies, which must first be inferred in order to conduct meaningful network analytics~\cite[Ch. 7]{kolaczyk2009statistical}; see also~ \cite{rodriguez2011uncovering,myers2010convexity,gomez2010inferring}. Prominent among these are SEMs, a family of statistical approaches for causal (a.k.a., path) analysis in complex systems, with several applications specifically tailored to graph topology inference; see e.g.,~\cite{goldberger1972structural,BGG14,cai2013inference}. In a nutshell, SEMs capture the relationship between observed nodal processes or measurements, and the  unknown causal network. The key contribution of SEMs is two-fold: a) they are conceptually simple, often resorting to tractable linear models; and b) SEMs explicitly account for the role played by exogenous or confounding inputs in observed nodal processes, which turn out to be critical in resolving directional ambiguities~\cite{bazerque2013identifiability}. 

In settings where measurement of exogenous inputs is costly or impractical, contemporary SEMs are quite limited with regard to unique identification of hidden network topologies. For example, in financial networks comprising stocks as nodes and their interdependencies as links, publicly-traded stock prices (endogenous) are known to depend on stock purchases (exogenous) by investors, whose details are often unknown to the public for privacy reasons. On the other hand, each publicly-traded company may broadcast monthly statistical summaries of purchases of its stock. Assuming that such statistical information is known or obtainable, the present paper advocates novel approaches that capitalize on factorization of carefully constructed \emph{tensors}, or multi-modal arrays. As demonstrated later, inference of the network topology is shown possible under reasonable conditions, using only correlation information of the exogenous inputs. The crux of our novel framework lies in positing that exogenous inputs exhibit piecewise-stationary correlations, from which three-way tensors are constructed using a special instance of SEMs. 

By leveraging the well-known parallel factor (PARAFAC) tensor decomposition~\cite{kolda2009tensor}, it is shown that edge connectivity information is captured through one of the factors, while identifiability of the network topology is guaranteed due to uniqueness of the factorization. Interestingly, casting the problem as tensor decomposition also opens up opportunities to {\it blindly }estimate both the unknown topology and \emph{local} correlation matrices of the exogenous inputs; see also~\cite{shen2016tensor,shen2016tracking}. PARAFAC decomposition is a powerful tool for multilinear algebra introduced by~\cite{harshman1970foundations}, and its merits have been permeated within diverse application domains \cite{sidiropoulos2016tensor}, e.g., wireless communications~\cite{sidiropoulos2000blind}, blind source separation~\cite{lee2013khatri,nion2010batch}, as well as community detection on graphs\cite{anandkumar2014tensor,papalexakis2013more}. The present paper broadens these well-documented merits to tasks involving network topology inference. Numerical tests on simulated and real data corroborate the efficacy of the novel approach.

Since most real-world networks are time-varying, the advocated tensor-based approach is accordingly extended to track topology changes. Moreover, nodal data are often acquired in real-time streams, rendering batch inference algorithms impractical. Toward satisfying the dual need to mitigate batch computational overhead, and track dynamic topologies, an online variant of the novel algorithm is developed. Motivated by the adaptive PARAFAC decomposition~\cite{nion2009adaptive,mardani2015subspace}, a novel real-time estimator is put forth to track the topology-revealing tensor factors, using second-order statistics of the exogenous inputs. 


To place this work in context, several prior studies have focused on tracking time-varying networks from nodal processes. For example, dynamic information diffusion networks were tracked via maximum likelihood estimators in \cite{gomez2013structure}, while a sparse piecewise stationary graphical model was put forth to track undirected networks in~\cite{angelosante2011sparse}. Dynamic SEMs were also advocated for inference of dynamic and directed cascade networks in~\cite{BGG14}. More recent work in \cite{tahani2016inferring} resorted to hidden Markov models (HMMs) to track diffusion links. 
 
PARAFAC decompositions have previously been advocated in e.g., blind source separation (BSS) tasks, which separate source signals from their mixed observations; see e.g., \cite{nion2010batch,lee2013khatri}. It is worth mentioning at the outset that tensor-based SEMs present unique challenges not encountered in traditional BSS, namely: i) network topologies are not directly revealed by factors obtained from the tensor decomposition, and one must exploit special properties inherent to SEMs; and ii) the  inherent scaling and permutation ambiguities are affordable compromises in BSS, but intolerable in the context of topology identification. Identifiability conditions developed in this paper aim to address these challenges. Tensor factorizations have also recently been adopted in network analytics and graph mining. For instance, several community detection approaches leverage the flexibility of tensors to capture more complex connectivity patterns such as \emph{cliques} and \emph{egonets}; see e.g.,~\cite{anandkumar2014tensor,benson2015tensor}, and~\cite{sheikholeslami2016egonet}.

The rest of this paper is organized as follows. Preliminaries and a formal statement of the problem are given in Section~\ref{sec:pre}, while Section~\ref{sec:tensor} casts the problem as a tensor factorization. Section~\ref{sec:identifiability} presents identifiability results for the proposed framework, while a topology tracking algorithm is developed in Section~\ref{sec:tracking}. Finally, results of corroborating numerical tests on both synthetic and real data are presented in Section~\ref{sec:test}, while concluding remarks and a discussion of ongoing and future directions are given in Section~\ref{sec:conc}.

\textit{Notation}. Bold uppercase (lowercase) letters will denote matrices (column vectors), while operators $(\cdot)^{\top}$, $\lambda_{\max}(\cdot)$, will stand for matrix transposition, and maximum eigenvalue, respectively. The identity matrix will be denoted by $\mathbf{I}$, while $\ell_p$ and Frobenius norms will be denoted by $\|.\|_p$ and $\|.\|_F$, respectively. The operator $\vec(.)$ will vertically stack columns of its matrix argument, to form a vector. Finally, $ \mathbf{A} \otimes  \mathbf{B}$ will denote the Kronecker product of matrices $\mathbf{A}$ and $\mathbf{B}$, while $ \mathbf{A} \odot \mathbf{B}$ will denote their Khatri-Rao product, namely, $\mathbf{A} \odot \mathbf{B} := [\mathbf{a}_1 \otimes \mathbf{b}_1, \dots \mathbf{a}_N \otimes \mathbf{b}_N]$, where $\mathbf{A} := [\mathbf{a}_1, \dots, \mathbf{a}_N]$ and $\mathbf{B} := [\mathbf{b}_1, \dots, \mathbf{b}_N]$.

\section{Preliminaries and Problem Statement}\label{sec:pre}
\begin{figure*}
	\centering
	\includegraphics[width=10cm]{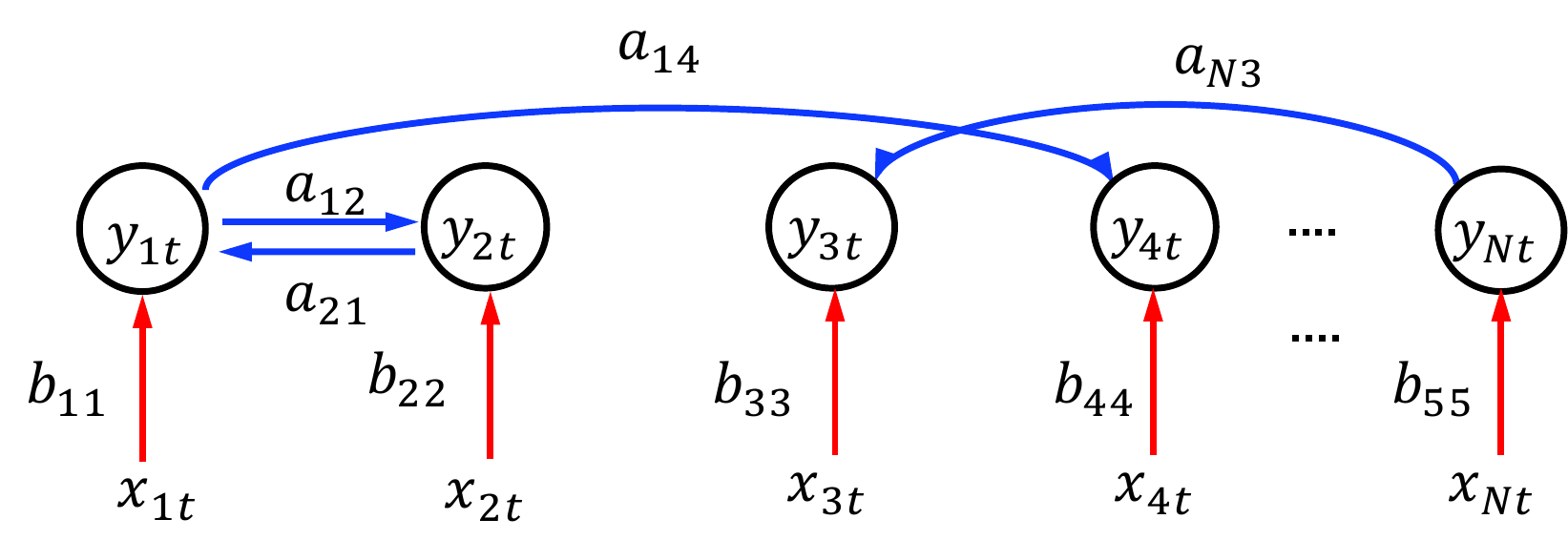}
	\caption{An $N$-node directed network (blue links), with the $t$-th samples of endogenous measurements per
node. SEMs explicitly account for exogenous inputs (red arrows), upon which endogenous variables may depend, in addition
to the underlying topology.}\label{fig:graph}
\end{figure*}

Consider a network $\mathcal{G}(\mathcal{V},\mathcal{E})$ that comprises $N$ nodes, with its topology captured by an unknown adjacency matrix $\bbA\in\mathbb{R}^{N\times N}$. Let $a_{ij}$ denote entry $(i,j)$ of $\bbA$, which is nonzero only if there is an edge between nodes $i$ and $j$; see Figure~\ref{fig:graph}. It will generally be assumed that $\mathcal{G}$ is a directed graph, that is $\bbA$ is a non-symmetric matrix $(\bbA \neq \bbA^{\top})$. 

Suppose the network abstracts a complex system with measurable inputs and an observable output process that propagates over the network following directed links. Let $x_{it}$ denote the input to node $i$ at slot $t$, and $y_{it}$ the $t$-th observation of the propagating process measured at node $i$. In the context of brain networks, $y_{it}$ could represent the $t$-th time sample of an electroencephalogram (EEG), or functional magnetic resonance imaging (fMRI) measurement at region $i$, while $x_{it}$ could be a controlled stimulus that affects a specific region of the brain. In social networks (e.g., \emph{Twitter} or \emph{Facebook}) over which information diffuses, $y_{it}$ could represent the timestamp when subscriber $i$ tweeted or shared a viral story, while $x_{it}$ could measure their level of interest in such stories.

In general, SEMs postulate that $y_{it}$ depends on two classes of variables, namely: i) measurements of the diffusing process $\{ y_{jt} \}_{j \neq i}$  (a.k.a. \emph{endogenous} variables); and ii) external inputs $x_{it} $ (a.k.a. \emph{exogenous} variables). Most contemporary SEM approaches posit that $y_{it}$ depends linearly on both $\{ y_{jt} \}_{j \neq i}$ and $ x_{it} $; that is,
 \begin{align}
 	y_{it}= \underbrace{\sum_{j\neq i}a_{ij}y_{jt}}_{\text{endogenous term}} + \underbrace{b_{ii} x_{it} }_{\text{exogenous term}}+e_{it}
	\label{mod:sem:linear}
 \end{align}
where $[\bbA]_{ij} := a_{ij}$, and $e_{it}$ denotes an ``error'' term that captures unmodeled dynamics. The coefficients $\{ a_{ij} \}$ and $\{ b_{ii} \}$ are unknown, and $a_{ij}\neq 0$ signifies that a directed edge from $j$ to $i$ is present. Collecting nodal measurements $\bby_t{ := }[y_{1t} \ldots y_{Nt}]^\top $, and $\bbx_t{ := }[x_{1t} \ldots x_{Nt}]^\top$ per slot $t$, and temporarily assuming that $e_{jt}=0$, the noise-free version of~\eqref{mod:sem:linear} can be compactly written as
 \begin{align}
 \label{mod:vec}
 	{\bby}_t=\bbA{\bby}_t+\bbB\bbx_t
 \end{align} 
where $[\bbA]_{ii} = 0$ and $\bbB := \text{Diag}(b_{11}, \dots, b_{NN})$ denotes a diagonal coefficient matrix. 

Note that with $\bbB$ diagonal,~\eqref{mod:sem:linear} implicitly assumes that each node is associated with a single exogenous input. In fact, it is possible to generalize~\eqref{mod:sem:linear} to settings where a single exogenous input may be applied to several nodes, or where a single node may be the recipient of multiple inputs. This amounts to relaxing the restriction on $\bbB$, allowing it to take values from the set of non-diagonal square matrices. In addition, in more general SEMs $\bbx_t$ and $\bby_t$ are indirectly observed latent variables, each adhering to \emph{measurement} models, namely $\mathbf{u}_{yt} = \mathbf{C}_y {\bby}_t + \boldsymbol{\delta}_{yt} $ and $\mathbf{u}_{xt} = \mathbf{C}_x {\bbx}_t + \boldsymbol{\delta}_{xt} $, with corresponding noise terms $\boldsymbol{\delta}_{yt} $ and $\boldsymbol{\delta}_{xt} $; see e.g., \cite{kaplan09} for details. In this case, the noisy version $({\bby}_t=\bbA{\bby}_t+\bbB\bbx_t + \mathbf{e}_t)$ of~\eqref{mod:vec} is often referred to as the \emph{structural model}. This paper deals with settings where ${\bbx}_t$ and ${\bby}_t$ are directly observable, and there is no extra measurement model. The problem statement can now be formally stated as follows.

\noindent{\textbf{Problem statement:}} Given $\{ \bby_t, \bbx_t \}_{t=1}^T$, the goal is to recover the underlying directed network topology $\bbA$. 

\section{A Tensor Factorization Approach}
\label{sec:tensor}


%

Building upon~\eqref{mod:sem:linear}, this section puts forth a novel tensor factorization approach to unveil the hidden network topology. To this end, the following assumptions are adopted.

\noindent\textbf{(as0)} Exogenous data $\{ \bbx_t^{(m)} \}$ are \emph{piecewise-stationary} over time segments $t\in[\tau_{m}, \tau_{m+1} - 1], m=1,\dots,M+1$, each with a fixed correlation matrix $\bbR_m^x := \mathbb{E}\{ \bbx_t^{(m)} (\bbx_t^{(m)})^{\top} \}$;

\noindent\textbf{(as1)} Entries of $\bbx_t$ are zero mean and uncorrelated per $t$; that is, $\mathbb{E} \{ x_{it} x_{jt} \} = 0, \forall i \neq j$;

\noindent\textbf{(as2)} Matrix $(\bbI - \bbA)$ is invertible; and

\noindent\textbf{(as3)} Matrix $\bbB$ is diagonal with nonzero diagonal entries.

\noindent Under (as0) and (as2), it is possible to rewrite~\eqref{mod:vec} as
\begin{align}
\label{eq_tensor_1a}
\bby_t=(\bbI-\bbA)^{-1}\bbB\bbx_t =\boldsymbol{\mathcal{A}} \bbx_t
\end{align}
where $\boldsymbol{\mathcal{A}} := (\bbI-\bbA)^{-1}\bbB$, and superscript ${(m)}$ has been dropped with the understanding that $t$ stays within one segment, and thus \eqref{eq_tensor_1a} holds $\forall m$. The \emph{per segment} correlation matrix $\bbR_m^y := \mathbb{E}\{ \bby_t \bby_t^{\top} \} $ is thus given by (cf. \eqref{eq_tensor_1a})
\begin{align}
\label{eq_tensor_1b}
\bbR_m^y  = \boldsymbol{\mathcal{A}} \bbR_m^x \boldsymbol{\mathcal{A}}^{\top}, ~~ t \in [\tau_{m}, \tau_{m+1} - 1]. 
\end{align}
Under (as1), one can express~\eqref{eq_tensor_1b} as the weighted sum of rank-one matrices as
\begin{align}\label{eq:R}
	\bbR_m^y = \boldsymbol{\mathcal{A}} \text{Diag}(\boldsymbol{\rho}^x_m) \boldsymbol{\mathcal{A}}^\top =\sum_{i=1}^N \rho^x_{mi}\bbalpha_i\bbalpha_i^\top
\end{align}
where $\bbalpha_i$ denotes the $i$th column of $\boldsymbol{\mathcal{A}}$, and $\boldsymbol{\rho}^x_m := [\rho^x_{m1} \ldots \rho^x_{mN}]^\top$, with $\rho^x_{mi} :=\mathbb{E}(x_{it}^2)$, for $t\in  [\tau_{m}, \tau_{m+1} - 1]$. 

\begin{figure*}[tpb!]
\centering
\includegraphics[scale=0.4]{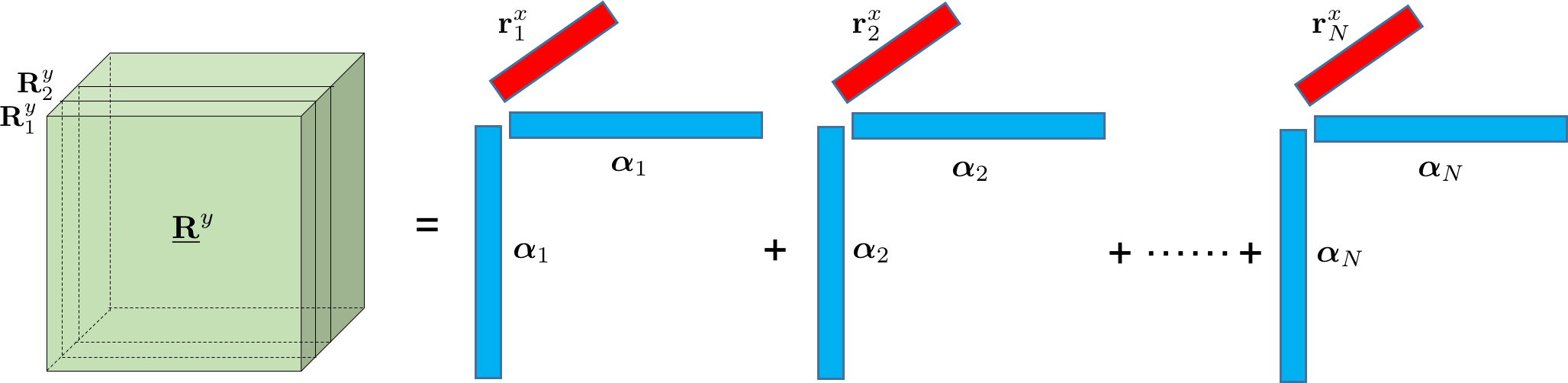}
\caption{The tensor $\underbar{\bbR}^y \in \mathbb{R}^{N \times N \times M}$ constructed by stacking the correlation matrices $\{ \bbR_m^y \in \mathbb{R}^{N \times N} \}_{m=1}^M$ admits a PARAFAC decomposition comprising rank-one tensor outer products. }
\label{fig:tensor}
\end{figure*}

Consider the three-way tensor $\underbar{\bbR}^y \in \mathbb{R}^{N \times N \times M}$, constructed by setting the $m$-th slice $\left[\underbar{\bbR}^y\right]_{:,:,m}=\bbR_m^y$. Letting $\alpha_{ji} \beta_{ki} \gamma_{li}$ denote the $(j,k,l)$ entry of the tensor outer product $\boldsymbol{\alpha}_i \circ \boldsymbol{\beta}_i \circ \boldsymbol{\gamma}_i$, where $\alpha_{ji} := [\boldsymbol{\alpha}_i]_j$ (resp. $\beta_{ik}$ and $\gamma_{il}$), it turns out that $\underbar{\bbR}^y$ can be written as (see also Figure~\ref{fig:tensor})
\begin{align}
\label{eq:tensor}
	\underbar{\bbR}^y=\sum_{i=1}^N \bbalpha_i\circ \bbalpha_i\circ \bbr^x_i
\end{align}
with entry $(j,k,l)$ given by 
\begin{align}
\label{eq:tensor_a}
	\left[ \underbar{\bbR}^y \right]_{jkl} = \sum_{i=1}^N \alpha_{ji}  \alpha_{ki} r^x_{li}
\end{align}
where $\bbr^x_i := [\rho^x_{1i} \dots \rho^x_{Mi}]^{\top}$. Interestingly,~\eqref{eq:tensor} amounts to the so-termed partial symmetric PARAFAC decomposition of $\underbar{\bbR}^y$ into factor matrices $\boldsymbol{\mathcal{A}}$, $\boldsymbol{\mathcal{A}}$, and $\bbR^x := [\bbr^x_1\dots \bbr^x_N] \in\mathbb{R}^{M\times N}$; see e.g., \cite{kolda2009tensor}.  
Although $\bbR_m^y$ is generally unknown, it can be readily estimated using sample averaging as  
\begin{align}
\label{eq:Rhat}
	\widehat{\bbR}^y_{m} = \frac{1}{\tau_{m+1} - \tau_m} \sum_{t = \tau_m}^{\tau_{m+1}-1} \bby_t \bby_t^{\top}, \quad m=1,\ldots,M
\end{align}
from endogenous measurements. 

The present paper relies on this three-way tensor constructed from second-order statistics of the nodal measurements, and leverages the uniqueness properties inherent to PARAFAC decompositions to identify the hidden network topology; see e.g.,~\cite{kruskal1977three} for key uniqueness results. Indeed, a number of standard PARAFAC decomposition algorithms can be adopted to estimate $\boldsymbol{\mathcal{A}}$; e.g., via alternating least-squares (ALS) iterations. Under reasonable conditions, it will be possible to recover $\bbA$, once $\boldsymbol{\mathcal{A}}$ has been found. The next proposition formally states the sufficient conditions required to uniquely identify $\bbA$, after determing of $\boldsymbol{\mathcal{A}}$ from the PARAFAC decomposition.

\begin{proposition}
\label{proposition1}
If $a_{jj} = 0$, $b_{jj} \neq 0 \; \forall j$, $b_{ij} = 0 \; \forall i \neq j$, and $\boldsymbol{\mathcal{A}}$ is invertible, then $\bbA$ can be uniquely expressed in terms of $\boldsymbol{\mathcal{A}}$ as $\bbA =\bbI-\left(\text{Diag}(\boldsymbol{\mathcal{A}}^{-1})\right)^{-1}\boldsymbol{\mathcal{A}}^{-1}$. 
\end{proposition}

\noindent \textbf{Proof:} See Appendix~\ref{appendix2}.
%

Regarding the decomposition in~\eqref{eq:tensor}, one can make the following important observations: 
(i) $\text{rank}(\underbar{\bbR}^y) = N$; (ii) two factors of $\underbar{\bbR}^y$ are identical;
and (iii) exogenous inputs $\{ \bbx_t \}_{t=1}^T$ are generally accessible, and can be readily tuned to satisfy piecewise stationarity along with the additional conditions necessary to guarantee identifiability of $\boldsymbol{\mathcal{A}}$. 

To quantify accessibility in (iii), one can consider $\bbR^x_{\Omega}$ known a priori, where $\Omega$ denotes the index set of the available entries of $\bbR^x$, i.e., $[\bbR^x_{\Omega}]_{i,j}=r^x_{ij}$ for $(i, j)\in \Omega$. Given noisy tensor data, these considerations (i)--(iii) prompt the next criterion for obtaining the wanted factors
\begin{align}
	(\hat{\bbZ}_1, \hat{\bbZ}_2, \hat{\bbZ}_3)&= \arg\min_{\bbZ_1, \bbZ_2, \bbZ_3} \bigg \| \underbar{\bbR}^y-\sum_{n=1}^N \bbz_{1n}\circ\bbz_{2n}\circ\bbz_{3n} \bigg \|_F^2 \nonumber\\
	\text{s.t.  } \quad  \bbZ_1&=\bbZ_2, ~~ [\bbZ_3]_{i,j}=\left[\bbR^x_{\Omega}\right]_{i,j},~~ \forall (i,j)\in\Omega\qquad\text{(P1)}\nonumber
\end{align}
where $\bbz_{in}$ denotes the $n$-th column of matrix $\bbZ_i$. Note that (P1) can be solved via partially symmetric PARAFAC decomposition, even when noise is present, using e.g., the individual differences in multidimensional scaling~\cite{carroll1970analysis}. Upon obtaining the estimated factors $\hat{\bbZ}_1, \hat{\bbZ}_2$ and $\hat{\bbZ}_3$, matrix $\hat{\bbA}$  can be found as (cf. Proposition \ref{proposition1})
\begin{eqnarray}
\widehat{\boldsymbol{\mathcal{A}}}&=&\hat{\bbZ}_1\\
		\hat{\bbA} &=&\bbI-\left(\text{Diag}(\widehat{\boldsymbol{\mathcal{A}}}^{-1})\right)^{-1}\widehat{\boldsymbol{\mathcal{A}}}^{-1}.
\end{eqnarray}

Unlike~\cite{bazerque2013identifiability} where explicit knowledge of the exogenous inputs is assumed to ensure model identifiability, our novel approach here establishes that knowledge of the second-order statistics captured through $\bbR^x$ could suffice. Detailed conditions under which the novel approach uniquely identifies the topology will be provided in Section \ref{sec:identifiability}. Algorithm~\ref{algo:netid} summarizes the resulting network topology inference scheme. It is assumed that one is given endogenous measurements $\{ \bby_t \}_{t=1}^T$, as well as $\bbR^x_{\Omega}$. It is also worth pointing out that S1 constructs $\underbar{\bbR}^y$ from endogenous data using the sample correlation matrices in \eqref{eq:Rhat}, since local correlation matrices $\{ \bbR_m^y\}_{m=1}^M$ are not explicitly known. The prescribed threshold $\eta$ in S4 is employed to determine the presence of edges. Its selection will be discussed in Section \ref{sec:test}.

\begin{remark}
The PARAFAC decomposition generally assumes no prior knowledge about $\bbR^x$; that is, $\Omega=\varnothing$ in (P1). In principle, one can estimate the topology even without correlation information of the exogenous inputs. Interestingly, this amounts to blindly estimating the topology and exogenous correlation matrices, which is of considerable merit when measurement of external inputs is impossible, or rather costly. 
\end{remark}

\begin{algorithm}[t] 
	\caption{Topology inference via tensor decomposition
	}\label{algo:netid}
	\begin{algorithmic} 
		\State\textbf{Input:}~$\bbR^x_{\Omega}$, $\{\bby_t\}$, $M$, $\eta$

		\State {\bf S1.} Tensor construction:\\ 
		\hspace{1.cm}Set $m$-th frontal slice of $\underbar{\bbR}^y \in \mathbb{R}^{N \times N \times M}$ to
		\vspace{1mm} 
		\\ \hspace{1.cm}$\widehat{\bbR}^{y}_m = \frac{1}{\tau_{m+1} - \tau_m} \sum_{t = \tau_m}^{\tau_{m+1}-1} \bby_t \bby_t^{\top}, ~m=1, \ldots, M$
			\vspace{1.5mm}

		\State {\bf S2.} PARAFAC decomposition:\\ 
		\vspace{1mm}
		\hspace{1.cm} Solve (P1) to find $(\hat{\bbZ}_1,\hat{\bbZ}_2,\hat{\bbZ}_3)$	
		
		\vspace{1.5mm}
		\State {\bf S3.} SEM estimates for topology inference:\\
		\vspace{1mm}
		\hspace{1.cm}$\widehat{\boldsymbol{\mathcal{A}}}=\hat{\bbZ}_1$\\
		\hspace{1.cm}$\hat{\bbA} =\bbI-\left(\text{Diag}(\widehat{\boldsymbol{\mathcal{A}}}^{-1})\right)^{-1}\widehat{\boldsymbol{\mathcal{A}}}^{-1}$ 
		\State  {\bf S4.} Edge identification:\\
		\hspace{1.cm}$[\hat{\bbA}]_{ij}\neq 0$ if $[\hat{\bbA}]_{ij}>\eta$, otherwise $[\hat{\bbA}]_{ij}= 0$, $\forall (i,j)$
		
	\end{algorithmic}
\end{algorithm}

\section{Identifiability issues}\label{sec:identifiability}
Although casting network topology identification task as a tensor decomposition problem leads to enhanced flexibility, one has to contend with identifiability issues common to both matrix and tensor factorizations. In order to establish identifiability conditions for $\bbA$ and $\bbB$, this section will first  explore conditions under which  $\boldsymbol{\mathcal{A}}$ is uniquely identifiable. To this end, a couple of definitions are in order.

\noindent\textbf{Definition 1.} The \emph{Kruskal rank} of a matrix $\mathbf{Z} \in \mathbb{R}^{N \times M}$ (denoted hereafter as $\text{kr}({\mathbf{Z}})$) is defined as the maximum number $k$ such that \emph{any} combination of $k$ columns of $\mathbf{Z}$ constitutes a full rank submatrix. 

\noindent \textbf{Definition 2.} \emph{Essential} uniqueness of a tensor factorization refers to uniqueness up to scaling and permutation ambiguity.

With Definitions $1$ and $2$ in mind, consider PARAFAC decomposition for a three way tensor $\underbar{\bbP}=(\bbU,\bbV,\bbW)$. Theorem~\ref{theorem:kruskal} establishes sufficient conditions for essential uniqueness of the tensor decomposition; see~\cite{stegeman2007kruskal} and \cite{kruskal1977three} for further details and a proof of the theorem.
\begin{theorem}
\label{theorem:kruskal}
\textit{Let $(\bbU,\bbV,\bbW)$ denote the PARAFAC factors obtained by decomposing
a three-way tensor $\underbar{\bbP}$ into $K$ rank-one tensors. If Kruskal's condition holds, namely,
\begin{align}
\label{cond:kruskal}
	{\rm kr}({\bbU})+{\rm kr}({\bbV})+{\rm kr}({\bbW})\geq 2K+2
\end{align}
and there exists an alternative set of matrices $ ( \bar{\bbU}, \bar{\bbV}, \bar{\bbW}) $ constituting a
PARAFAC decomposition of $\underbar{\bbP}$, then there exists a permutation matrix $\bbPi$, and diagonal scaling matrices 
$\bbLambda_1$,  $\bbLambda_2$,  $\bbLambda_3$, such that $\bbLambda_1 \bbLambda_2 \bbLambda_3 = \bbI$,
$ \bar{\bbU}=\bbU\bbPi\bbLambda_1$ , $ \bar{\bbV}=\bbV\bbPi\bbLambda_2 $, and $\bar{\bbW}=\bbW\bbPi\bbLambda_3$. }
\end{theorem}

\noindent \textbf{Proof:} See~\cite{stegeman2007kruskal} for a general proof with complex entries.

As a prerequisite to identification of $\bbA$, the following proposition establishes essential uniqueness of $\boldsymbol{\mathcal{A}}$, 
based on the tensor-based interpretation advocated in the prequel.
%
%
%

\begin{proposition}
\label{theorem1}
If ${\rm kr}(\bbR^x)\geq 2$, then $\boldsymbol{\mathcal{A}}:=(\bbI-\bbA)^{-1}\bbB$ is uniquely identifiable up to a scaling and permutation ambiguity via
PARAFAC decomposition of $\underbar{\bbR}^y$.
\end{proposition}

\noindent \textbf{Proof:} Upon recognizing that $\text{rank}(\underbar{\bbR}^y) = N$ from~\eqref{eq:tensor}, in order for~\eqref{cond:kruskal} to hold, we need
\begin{align}
\label{cond1}
	2\text{kr}({\boldsymbol{\mathcal{A}}})+\text{kr}({\bbR^x})\geq 2N+2.
\end{align}
Under (as2) and (as3), matrices $(\bbI-\bbA)$ and $\bbB$ are invertible, which implies that $\boldsymbol{\mathcal{A}}=(\bbI-\bbA)^{-1}\bbB$ is invertible, and hence $\text{kr}({\boldsymbol{\mathcal{A}}})=N$. From \eqref{cond1}, essential uniqueness can thus be guaranteed as long as $\text{kr}({\bbR^x})\geq 2$, which completes the proof.
%

Note that essential uniqueness is not sufficient for identification of the hidden network topology, due to the inherent permutation and scaling ambiguities. To this end, we will subsequently pursue identifiability conditions for settings where $\bbR^x$ may be fully, or partially available, or even completely unavailable on a case-by-case basis.

\subsection{Identifiability with fully known $\bbR^x$}
\label{subsec:full_knowledge}
First, we will explore identifiability of the topology when $\bbR^x$ is completely known, while highlighting the importance of information about exogenous inputs $\{\bbx_t\}$.

\begin{theorem}
\label{theorem:full}
\textit{If $\bbx_{t}$ and $\bby_t$ obey the SEM in \eqref{mod:vec}, for all $t=1, \ldots$, with $\bbA$ and $\bbB$ satisfying (as2) and (as3), respectively, and if $\bbR^x$ is known and satisfies ${\rm kr}({\bbR^x})\geq 2$, then $\bbA$ can be uniquely identified via Algorithm \ref{algo:netid}.}
\end{theorem}

\noindent\textbf{Proof:} Suppose there is an alternative triplet $( \boldsymbol{\mathcal{A}}', \boldsymbol{\mathcal{A}}', {\bbR^x}')$, also decomposing $\underbar{\bbR}^y$ into $N$ rank-one tensors in (P1). Theorem \ref{theorem:kruskal} asserts that there is a permutation matrix $\bbPi$, and diagonal scaling matrices $\{ \bbLambda_1, \bbLambda_2, \bbLambda_3 \}$ so that
	\begin{align}
	\label{cond:lambda1}
		\bbLambda_1 \bbLambda_2 \bbLambda_3 = \bbI
	\end{align}
	 and
	\begin{subequations}
	\begin{eqnarray}
		\label{cond:Pi1}
		\boldsymbol{\mathcal{A}}'&=&\boldsymbol{\mathcal{A}}\bbPi\bbLambda_1\\
		\label{cond:Pi2}
		 \boldsymbol{\mathcal{A}}'&=&\boldsymbol{\mathcal{A}}\bbPi\bbLambda_2\\
		 \label{cond:Pi3}
		  {\bbR^x}'&=&\bbR^x\bbPi\bbLambda_3
	\end{eqnarray}	
	\end{subequations}
where one can readily deduce from~\eqref{cond:Pi1} and~\eqref{cond:Pi2} that $\bbLambda_1=\bbLambda_2$.
On the other hand, when $\bbR^x$ is known a priori, i.e., $\bbR_{\Omega}^x=\bbR^x$, the constraint in (P1) yields ${\bbR^x}'=\bbR^x$. Consequently, \eqref{cond:Pi3} can be written as
	\begin{align}
	\label{eq:lemmac1}
		 \bbR^x=&\bbR^x\bbPi\bbLambda_3
	\end{align}
for which the following holds.
	
\begin{lemma}\label{lemma1}
For permutation matrix $\bbPi$, scaling matrix $\bbLambda_3$, and $\bbR^x$ satisfying the inequality ${\rm kr}({\bbR^x})\geq 2$, \eqref{eq:lemmac1} holds true if and only if 
		\begin{subequations}
		\begin{eqnarray}
		\label{eq:lemma:pi}
				\label{eq:lemma:lambda}
			\bbLambda_3 &=& \bbI\\
				\label{eq:lemma:pi}
			\bbPi &=& \bbI .
		\end{eqnarray}
			
		\end{subequations}

\end{lemma}

\noindent\textbf{Proof:} See Appendix~\ref{app:lemmac}.

\noindent Next, substituting \eqref{eq:lemma:pi} into \eqref{cond:Pi1}, and letting $\bbLambda=\bbLambda_1=\bbLambda_2$, one obtains
\begin{align}
\label{eq:Phi}
	\boldsymbol{\mathcal{A}}'=\boldsymbol{\mathcal{A}}\bbLambda.
\end{align}
for which the next lemma holds true.

\begin{lemma}
\label{lemma2}
If the PARAFAC solution obtained in S3 of Algorithm~\ref{algo:netid} satisfies $\widehat{\boldsymbol{\mathcal{A}}}=\boldsymbol{\mathcal{A}}\bbLambda$, then $\bbA$ can be uniquely identified; that is, $\hat{\bbA}=\bbA$.
\end{lemma}

\noindent\textbf{Proof:} See Appendix~\ref{app:lemma2}.

\noindent Combining Lemma~\ref{lemma2} with~\eqref{eq:Phi} completes the proof of Theorem~\ref{theorem:full}.



\subsection{Identifiability with partially known $\bbR^x$}

The last subsection assumed that second-order statistics of $\bbx_t$ were available for all time slots $m=1, \ldots, M$. However, ample empirical evidence suggests that such information may not be fully available at times. For instance, not all statistics of the stock prices may be available to a given investor in financial markets over time. In brain connectivity studies, one may only have explicit knowledge about exogenous variables in some experimental settings, but not others. Such limitations motivate the analysis of identifiability in settings where one only has access to partial information about second-order statistics of exogenous inputs; that is, $\bbR^x$ contains misses.

In order to capture the partial availability of $\bbR^x$, suppose $\Omega_i$ denotes set of indices corresponding to known entries per column $i$ of $\bbR^x$. Furthermore, let $\check{\bbr}_i^j$ denote a sub-vector of $\bbr^x_i$, whose entries are indexed by $\Omega_i\cup\Omega_j$ (recall that 
$\bbr^x_i$ denotes the $i$-th column of $\bbR^x$). Based on these definitions, the next theorem establishes identifiability conditions for settings where $\bbR^x$ is only partially available.

\begin{theorem}
\label{theorem:semi_blind}
\textit{If $\check{\bbr}_i^j$ and $\check{\bbr}_j^i$ are linearly independent for any $i\neq j$, then the network adjacency matrix $\bbA$ can be uniquely identified via Algorithm \ref{algo:netid}.}
\end{theorem}

\noindent \textbf{Proof:}	 
Suppose there exists an alternative PARAFAC solution $( \check{\boldsymbol{\mathcal{A}}}, \check{\boldsymbol{\mathcal{A}}}, \check{\bbR}^x)$ that also decomposes $\underbar{\bbR}^y$ into $N$ rank-one tensors (cf. S2 in Algorithm~\ref{algo:netid}). According to Theorem~\ref{theorem:kruskal}, there exists a permutation matrix $\check{\bbPi}$ and diagonal scaling matrices $\{ \check{\bbLambda}_1, \check{\bbLambda}_2, \check{\bbLambda}_3 \}$ such that
	\begin{align}
	\label{cond:lambda1s}
		\check{\bbLambda}_1 \check{\bbLambda}_2 \check{\bbLambda}_3 = \bbI
	\end{align}
and
	\begin{subequations}
	\begin{eqnarray}
		\label{cond:Pi1s}
		\check{\boldsymbol{\mathcal{A}}}&=&\boldsymbol{\mathcal{A}}\check{\bbPi}\check{\bbLambda}_1\\
		\label{cond:Pi2s}
		 \check{\boldsymbol{\mathcal{A}}}&=&\boldsymbol{\mathcal{A}}\check{\bbPi}\check{\bbLambda}_2\\
		 \label{cond:Pi3s}
		  \check{\bbR}^x&=&\bbR^x\check{\bbPi}\check{\bbLambda}_3
	\end{eqnarray}	
	\end{subequations}
where from \eqref{cond:Pi1} and \eqref{cond:Pi2}, it is clear that $\check{\bbLambda}_1=\check{\bbLambda}_2$.
	%
On the other hand, when $\bbR^x$ is partially known;  that is, $[\check{\bbR}^x]_{i,j}=\left[\bbR^x\right]_{i,j}$, for $(i,j)\in\Omega$,
	%
	%
then~\eqref{cond:Pi3s} can be written as
	\begin{align}
	\label{eq:lemmac1s}
		 [\bbR^x]_{i,j}=&[\bbR^x\check{\bbPi}\check{\bbLambda}_3]_{i,j},\qquad \forall~~(i,j)\in\Omega.
	\end{align}
The rest of the proof of Theorem~\ref{theorem:semi_blind} builds on the following lemma.

\begin{lemma}\label{lemma3}
For a given permutation matrix $\check{\bbPi}$, and scaling matrix $\check{\boldsymbol{\Lambda}}_3$, if $\bbR^x$ satisfies the condition in Theorem~\ref{theorem:full}, then \eqref{eq:lemmac1s} holds true if and only if 
\begin{subequations}
\begin{eqnarray}
\label{eq:lemma:pis}
			\check{\bbPi} &=&\bbI\\
			\label{eq:lemma:lambdas}
			\check{\bbLambda}_3 &=& \bbI.	
		\end{eqnarray}					
\end{subequations}
\end{lemma}
\noindent \textbf{Proof:} See Appendix \ref{app:lemmac:s}.

\noindent Upon substituting of~\eqref{eq:lemma:pis} into~\eqref{cond:Pi1s}, and letting $\check{\bbLambda}=\check{\bbLambda}_1=\check{\bbLambda}_2$, it turns out that
	\begin{align}
		\check{\boldsymbol{\mathcal{A}}}=\boldsymbol{\mathcal{A}}\check{\bbLambda}
	\end{align}
and the conclusion of Theorem~\ref{theorem:full} follows from Lemma \ref{lemma2}.
\begin{remark}
The central premise of Theorem \ref{theorem:full} is that even when $\bbR^x$ contains misses, it is possible to uniquely identify the adjacency matrix $\bbA$. In turn, this facilitates the combination of information pertaining to nodal processes from different time slots towards the task of inference of the hidden network topology, even though complete correlation information is unavailable for all the nodes.
\end{remark}

Our novel tensor-based topology identification approach advocated so far focuses on settings where the network topology does not vary with time. The rest of the paper goes beyond this assumption, and explores scenarios where the link structure may even evolve over time, with the ultimate goal of tracking the network topology, possibly in real time.

\section{Tracking dynamic network topologies}
\label{sec:tracking}

It has hitherto been taken for granted that all past data are available, and the developed tensor-based approaches will operate in batch mode. In fact,  Algorithm~\ref{algo:netid} is conducted entirely offline, with $\underbar{\bbR}^y$ obtained or computed a priori. However, practical constraints often render it impossible to operate in batch mode; for instance, nodal data in large-scale networks (e.g., modern social media and the web) can only be acquired in real-time streams since any attempts to store such data for batch processing will quickly overwhelm operators. 

Equally important is the observation that most real-world networks evolve over time, namely, new edges and nodes may appear, while others become obsolete during the observation period. Consequently, even if a batch approach were to overcome challenges due to the sheer scale of the data, the inferred networks would represent a single aggregate perspective of several evolving network topologies at best. In lieu of these challenges, this section extends the novel tensor-based approach to track changes to the network topologies in real time. 

\subsection{Piecewise-invariant dynamic network topologies}
\label{subsec:piecewise}

Suppose that the network exhibits a piecewise-constant topology, captured by the sequence of unknown adjacency matrices $\{ \bbA_m \in \mathbb{R}^{N\times N}, \; t \in [\tau_m, \tau_{m+1} - 1] \}_{m=1}^M$, over $M$ time segments. Each entry $(i,j)$ of $\bbA_m$ is nonzero only if a directed edge exists from node $i$ to $j$, and it will be denoted by $a_{ij}^m$. Similarly associating each node with a single exogenous input, one obtains the following SEM
 \begin{align}
 	y_{jt}=\sum_{i\neq j}a_{ij}^my_{it} +b_{jj}^m x_{jt} +e_{jt}, \;\;\; t \in [\tau_m, \tau_{m+1} - 1]
	\label{mod:sem:linear:dynamic}
 \end{align}
per $m = 1,\dots, M$, with $e_{jt}$ similarly capturing unmodeled dynamics, while coefficients $\{ a_{ij}^m \}$ and $\{ b_{jj}^m \}$ are unknown. 
With $\bby_t$, $\bbx_t$, and $\mathbf{e}_t $ previously defined,~\eqref{mod:sem:linear:dynamic} can be written in vector form as
 \begin{align}
 \label{mod:vec:dynamic}
 	{\bby}_t=\bbA_m{\bby}_t+\bbB_m\bbx_t + \mathbf{e}_t
 \end{align} 
where $[\bbA_m]_{ij} = a_{ij}^m$ and $\bbB_m := \text{Diag}(b_{11}^m, \dots, b_{NN}^m)$. Based on \eqref{mod:vec:dynamic}, we will develop an algorithm to track $\{ \bbA_m, \bbB_m \}_{m=1}^M$ using measured endogenous variables, and the sequence of correlation matrices $\{ \bbR_m^x \}_{m=1}^M$.

\begin{figure}[t!]
\centering
\includegraphics[scale=0.38]{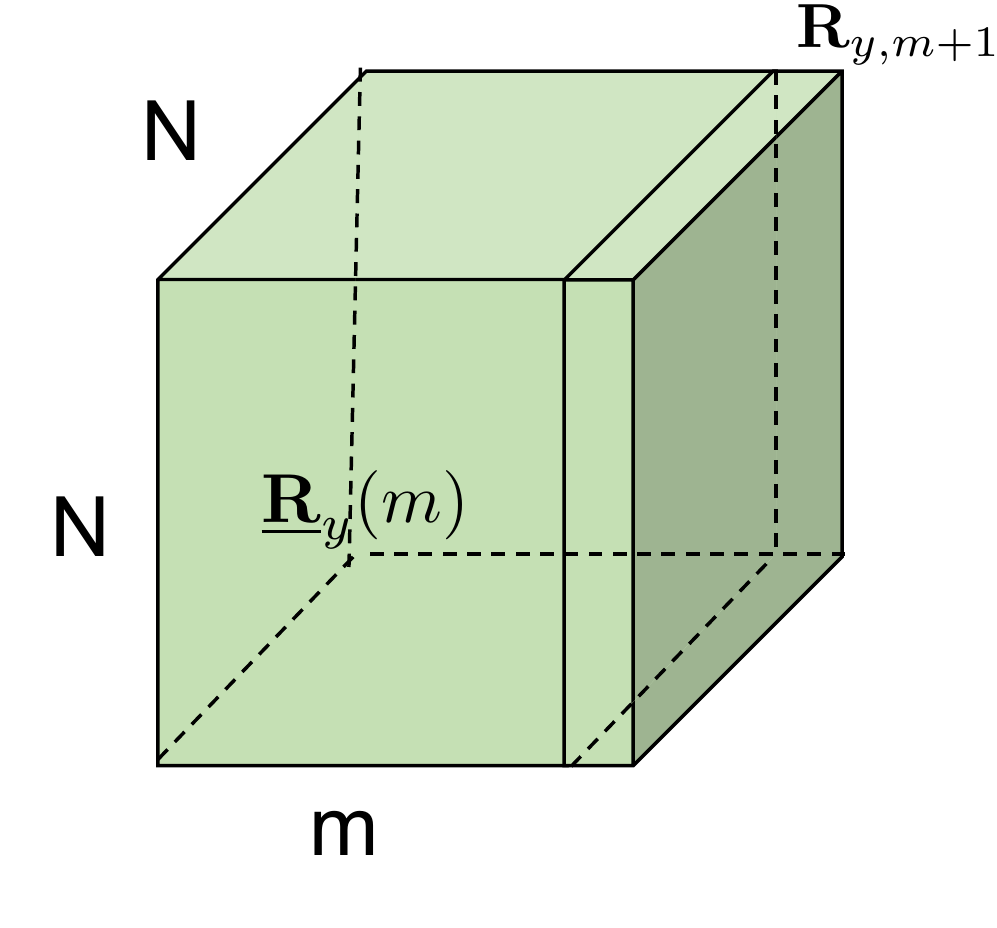}
\caption{Tensor $\underbar{\bbR}^y$ grows per window $m$ by a new frontal slice  $\bbR_m^y \in \mathbb{R}^{N \times N} $. }
\label{fig:tensor_dynamic}
\end{figure}

Key to the novel topology tracking algorithm is recognizing that the tensor-based approach of Section \ref{sec:tensor} can be extended to settings where the network exhibits piecewise-constant topology variations. To this end, define $\boldsymbol{\mathcal{A}}_m:=(\bbI-\bbA_m)^{-1}\bbB_m$, and consider a tensor with the $m$-th slice
\begin{equation}
\label{eq:track0}
	\bbR_m^y 	= \boldsymbol{\mathcal{A}}_m \bbR_m^x \boldsymbol{\mathcal{A}}_m^{\top}, \quad t \in [\tau_{m}, \tau_{m+1} - 1] 
\end{equation}
sequentially appended at $t=\tau_{m+1}$, for $m = 1, \dots, M$; see also~\eqref{eq:R} and Figure~\ref{fig:tensor_dynamic}. Allowing $\underbar{\bbR}^y$ to grow sequentially along one mode is well motivated for real-time operation, where data may be acquired in a streaming manner. In this case, unveiling the evolving network topology calls for approaches that are capable of tracking tensor factors. In fact, the topology tracking algorithm developed next builds upon a prior sequential tensor factorization approach, namely, PARAFAC via recursive least-squares tracking (PARAFAC-RLST); see e.g., ~\cite{nion2009adaptive} for details. 

\subsection{Exponentially-weighted least-squares estimator}
\label{subsec:ewlse}

Let $\bar{\bbr}_m^y:=\vec(\bbR_m^y)$ denote the vectorization of $\bbR_m^y$, and note that $\bar{\bbr}_m^y$ can be written as $ \bar{\bbr}_m^y=\bbH_m\bbrho^x_m $, where $\bbH_m:=\boldsymbol{\mathcal{A}}_m\odot\boldsymbol{\mathcal{A}}_m$ is an $N^2\times N$ matrix, and $\boldsymbol{\rho}^x_m$ is defined after \eqref{eq:R}. To track $\bbH_m$, we advocate an exponentially-weighted least-squares estimator, namely,
\begin{align}
\label{prob:H}
	\widehat{\bbH}_m = \arg \min_{\bbH} \;\; \sum_{l=1}^m\beta^{m-l}\|\bar{\bbr}^y_{l}-\bbH\bbrho^x_l\|^2_2
\end{align}
for $m=1,\dots,M$, where $\beta\in(0,1]$ denotes a forgetting factor, which facilitates tracking topology changes by down-weighing past data when $\beta<1$. 

Letting  $f_m(\bbH) := \sum_{l=1}^m\beta^{m-l}\|\bar{\bbr}_l^y-\bbH\bbrho^x_l\|^2_2 $ denote the cost function per segment $m$, 
and taking the gradient with respect to $\bbH$, one obtains
\begin{align}
	\nabla f_m(\bbH)=2\sum_{l=1}^m\beta^{m-l}\left(\bar{\bbr}_l^y-\bbH\bbrho^x_l\right)(\bbrho^x_l)^\top.
\end{align}
Setting $\nabla f_m(\bbH) = \boldsymbol{0}$, and solving for $\bbH_m$ yields
\begin{align}
\label{eq:H}
	\bbH_m=\bbQ_m\bbP^{-1}_m
\end{align}
where $\bbQ_m:=\sum_{l=1}^m\beta^{m-l} \bar{\bbr}_l^y(\bbrho^x_l)^\top$ and 
$\bbP_m:=\sum_{l=1}^m\beta^{m-l}\bbrho^x_l(\bbrho^x_l)^\top$.
%
%
Further inspection of $\bbP_m$ and $\bbQ_m$ reveals that the updates admit recursive forms as follows
\begin{eqnarray}
\label{eq:P:update}
	\bbP_m &:=&\beta \bbP_{m-1}+\bbrho^x_m(\bbrho^x_m)^\top\\
\label{eq:Q:update}
	\bbQ_m &:=&\beta \bbQ_{m-1}+\bar{\bbr}_m^y(\bbrho^x_m)^\top.
\end{eqnarray}
Moreover, letting $\bbW_m:=\bbP^{-1}_m$, one can resort to the matrix inversion lemma to recursively 
compute inverses as

\begin{equation}
\label{eq:W:update}
	\bbW_m = \beta^{-1} \left[ \bbW_{m-1}
	 -\frac{\bbW_{m-1}\bbrho^x_m(\bbrho^x_m)^\top\bbW_{m-1}}{\beta+(\bbrho^x_m)^\top\bbW_{m-1}\bbrho^x_m} \right].
\end{equation}
It is worth pointing out that the simple recursive updates~\eqref{eq:P:update} 
-~\eqref{eq:W:update} lead to a markedly reduced computational burden, 
while only requiring fixed memory storage costs.


Once $\bbH_m$ is estimated, $\boldsymbol{\mathcal{A}}_m := [ \bbalpha_{1m}, \dots, \bbalpha_{Nm} ]$ can be recovered by recalling that the $i$th column of $\bbH_m$ is given by
\begin{align}
	\bbh_{im} =\bbalpha_{im}\otimes \bbalpha_{im} = \text{vec}(\bbalpha_{im}\bbalpha^\top_{im}).
\end{align}
Recognizing that $\bar{\bbH}_{im} := \bbalpha_{im} \bbalpha^\top_{im}$ is a rank one matrix, $\bbalpha_{im}$ can be estimated 
via the leading eigenvector of $\bar{\bbH}_{im}$, namely
\begin{align}
\label{phi:hat}
	\widehat{\bbalpha}_{im} \approx \lambda^{\frac{1}{2}}_{\text{max}}(\bar{\bbH}_{im}) \bbv_{\text{max}}(\bar{\bbH}_{im})
\end{align}
where the eigen-pair $\{ \lambda_{\text{max}}(\bar{\bbH}_{im}),  \bbv_{\text{max}}(\bar{\bbH}_{im})\}$ denotes the leading eigenvalue of $\bar{\bbH}_{im}$, and its corresponding eigenvector, both obtainable via the power iteration \cite{golub1996matrix}. This is carried out per column of $\boldsymbol{\mathcal{A}}_{m}$ to obtain 
$\widehat{\boldsymbol{\mathcal{A}}}_m := [ \widehat{\bbalpha}_{1m}, \dots, \widehat{\bbalpha}_{Nm} ]$, while $\bbA_m$ can be estimated as (cf. Proposition \ref{proposition1})
\begin{align}
\label{update:BA:online}
\hat{\bbA}_m = \bbI-\left(\text{Diag}(\widehat{\boldsymbol{\mathcal{A}}}^{-1}_m)\right)^{-1}\widehat{\boldsymbol{\mathcal{A}}}^{-1}_m.
\end{align}
Algorithm~\ref{algo:netid:online} lists the steps involved in tracking evolving network topologies via the scheme advocated in this section.

%
\begin{algorithm}[tpb!] 
	\caption{Tensor-based network topology tracking}\label{algo:netid:online}
	\begin{algorithmic} 
		\State\textbf{Input:}~$\{\bbrho^x_m\}_{m=1}^M$, $\{\bby_t\}$, $\beta$, $\bbW_0$, $\bbQ_0=\boldsymbol{0}$, $\eta$
		\For{$m=1, \ldots, M$}
		
		\State {\bf S1.} Tensor formation\\ 
		\hspace{1.3cm}Set frontal slice $m$ of $\underbar{\bbR}^y$  
		to $\widehat{\bbR}_{m}^y$ as in~\eqref{eq:Rhat} 			\vspace{1.5mm}

		\State {\bf S2.} Variable updates:\\ 
		\hspace{1.3cm}$\bbQ_m := \beta \bbQ_{m-1}+\bar{\bbr}_m^y(\bbrho^x_m)^\top$ \\
		\hspace{1.3cm}Update $\bbW_m$ via \eqref{eq:W:update}\\
		\hspace{1.3cm}Uptate $\widehat{\bbalpha}_{im}$ via \eqref{phi:hat}, for $i=1, \ldots N$
		\vspace{1.5mm}
		
		\State {\bf S3.} SEM estimates for topology tracking:\\
		\hspace{1.3cm}Estimate $\hat{\bbA}_m$ via \eqref{update:BA:online}.
		\State Return $\hat{\bbA}_m$
		\EndFor
		\State  \bf Edge identification:\\
		$[\hat{\bbA}_m]_{ij}\neq 0$ if $[\hat{\bbA}_m]_{ij}>\eta$, otherwise $[\hat{\bbA}_m]_{ij}= 0$, $\forall (i,j)$
	\end{algorithmic}
\end{algorithm}

\noindent\begin{remark}[Initialization]
Matrix $\bbP_m$ in \eqref{eq:P:update} is rank deficient when $m\leq N$, rendering the update in \eqref{eq:H} impossible. This can be addressed by setting $\bbW_0=\bbP_0^{-1}=a\bbI$, for a very large constant $a$ (e.g., $a = 10^5$). Since $\bbP^{-1}_m$ is a variance estimate of $\widehat{\bbH}_m$, this initialization amounts to placing little confidence in the initial values. Matrix $\bbQ_0$ is initialized as an all-zero matrix.
\end{remark}

 \section {Numerical Tests}
 \label{sec:test}

In order to assess the effectiveness of the novel algorithms, this section presents test results from experiments conducted on both simulated and real network data. Consideration was given to scenarios involving both static and dynamic networks.

\subsection{Tests on static simulated networks}
\label{subsec:sim_stat}
\noindent\textbf{Data generation.} A \emph{Kronecker random graph} comprising $N=64$ nodes was generated from a prescribed ``seed matrix'' 
\begin{align}
\bbS_0:=\left(
\begin{array}{cccc}
0 &	0&1&1\\
0 &	0&1&1\\
0 &	1&0&1\\
1 &	0&1&0
\end{array} \right) \nonumber
\end{align}
in order to obtain a binary-valued $64\times 64$ matrix via repeated Kronecker products, namely $\bbS=\bbS_0\otimes\bbS_0\otimes\bbS_0$; see also \cite{leskovec2010kronecker}. Using the binary matrix $\bbS$ to describe the zero and nonzero entries of the topology, the Kronecker graph with adjacency matrix $\bbA$ was then constructed by randomly sampling each entry from a uniform distribution with $a_{ij}\sim\text{Unif}(0.2s_{ij},0.5s_{ij})$. To generate synthetic endogenous measurements, the observation horizon was set to $T=ML$ time-slots, which were partitioned into $M$ windows of fixed length $L$, using pre-selected boundaries $\{ \tau_m \}_{m=1}^{M+1}$ with $\tau_1 = 1$ and $L := \tau_{m+1} - \tau_{m}$, for several values of $L$ and $M$. Per $t \in [\tau_{m}, \tau_{m+1} - 1]$, exogenous inputs were sampled as $\bbx_t \sim \mathcal{N}(\boldsymbol{0}, \sigma_m^2 \bbI)$, with $\{ \sigma_m \}_{m=1}^M$ set to $M$ distinct values. With $\mathbf{e}_t$ sampled i.i.d. from $\mathcal{N}(\mathbf{0}, 10^{-2} \bbI)$, $\bby_t$ was generated using the SEM, that is, $\bby_t = (\bbI - \bbA)^{-1}(\mathbf{B}\bbx_t + \mathbf{e}_t)$, where $\bbB$ is a diagonal matrix with $[\bbB]_{jj}$ drawn uniformly from the interval $[2, 3]$. 
 
In order to conduct PARAFAC decompositions, an implementation in the open source Tensorlab 3.0 toolbox was adopted~\cite{tensorlab3.0}. Upon running Algorithm~\ref{algo:netid}, an edge was declared present if the estimate $\hat{a}_{ij}$ was found to exceed a prescribed threshold. The threshold was selected to yield the lowest edge identification error rate (EIER), which is defined as
\begin{align}
\text{EIER}:=\frac{\|\bbS-\widehat{\bbS}\|_0}{N(N-1)}\times 100\%
\end{align}
with the operator $\|\cdot\|_0$ denoting the number of nonzero entries of its argument. Matrix $\bbS \in \{0,1\}^{N \times N}$ denotes the ground-truth binary edge indicator matrix, while $\widehat{\bbS}$ denotes its estimate obtained by the novel scheme. 
 
Experiments were run for different values of $M$, 
and error plots were generated using EIER values averaged over $500$ independent runs. 
 \begin{figure}[t]
	\centering
\includegraphics[width=8.7cm]{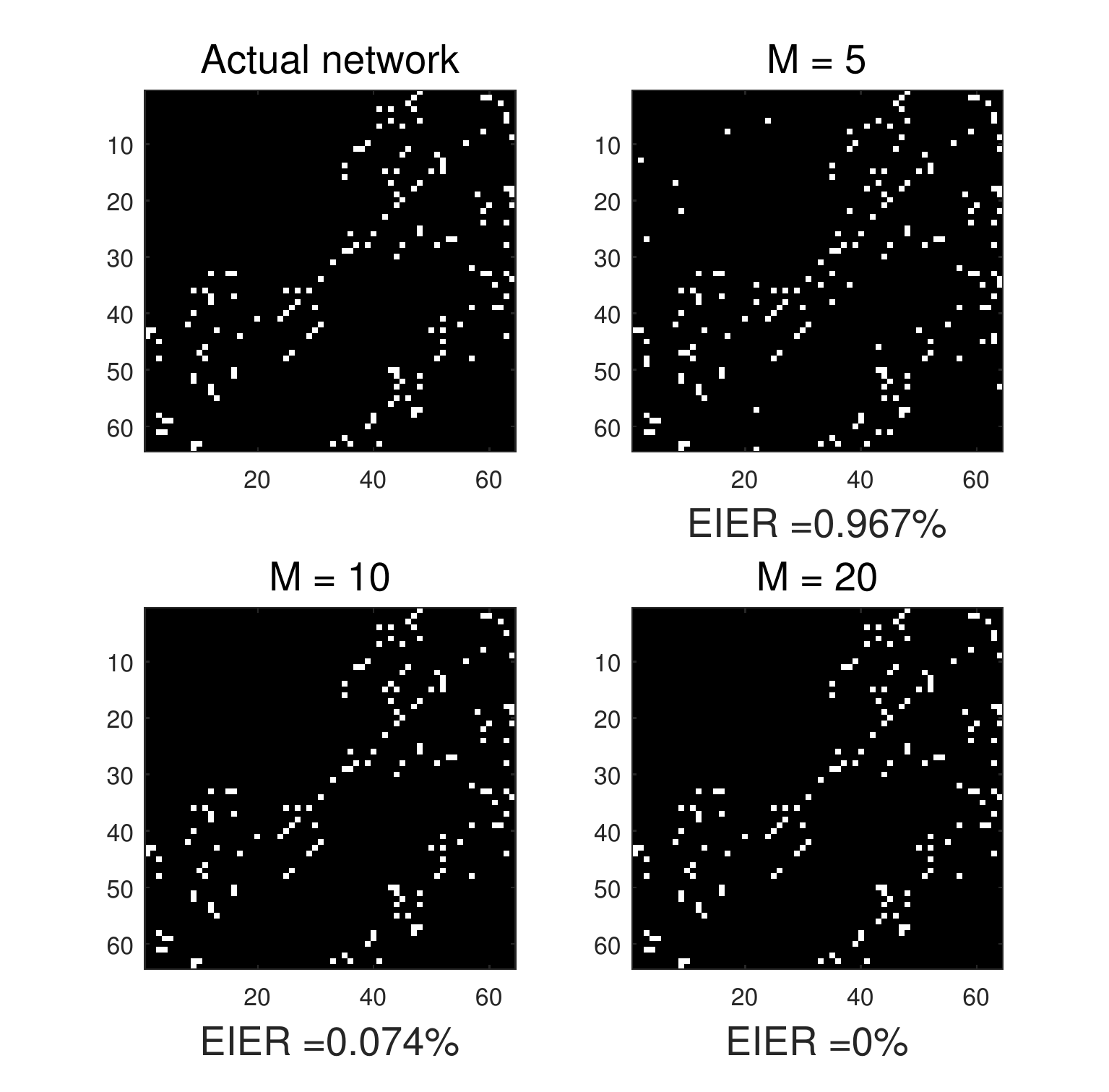}
\caption{Actual and inferred adjacency matrices with the number of windows set to $M=5, 10$, and $20$. }
\label{fig:heatmap}
\end{figure}
\begin{figure*}[tpb!]
\hspace{-1.5mm}
\begin{minipage}[b]{.32\textwidth}
\centering
\includegraphics[width=6cm]{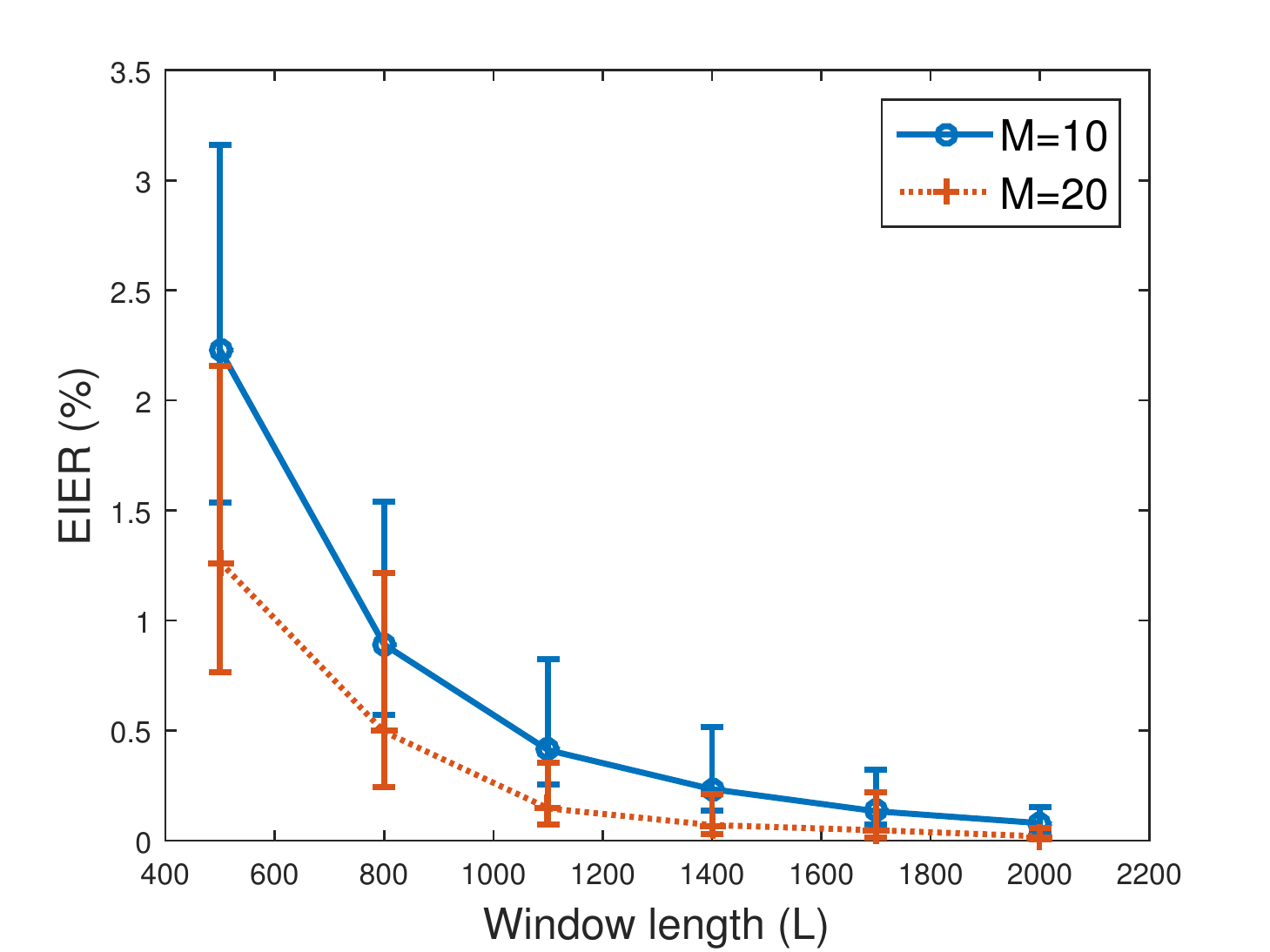}
\centerline{(a)}
\end{minipage}
\begin{minipage}[b]{.32\textwidth}
\centering
\includegraphics[width=6cm]{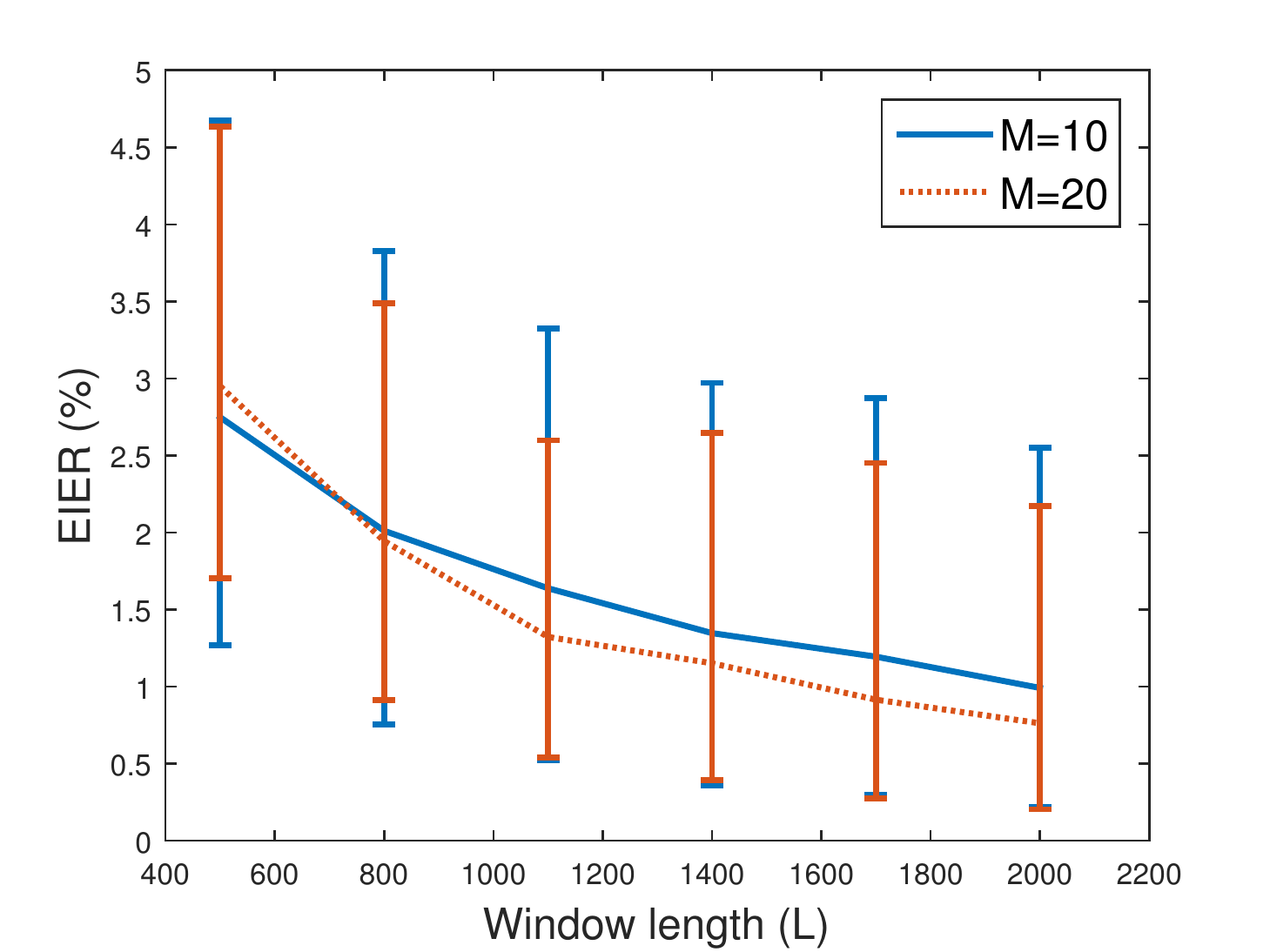}
\centerline{(b)}
\end{minipage}
\begin{minipage}[b]{.32\textwidth}
\centering
\includegraphics[width=6cm]{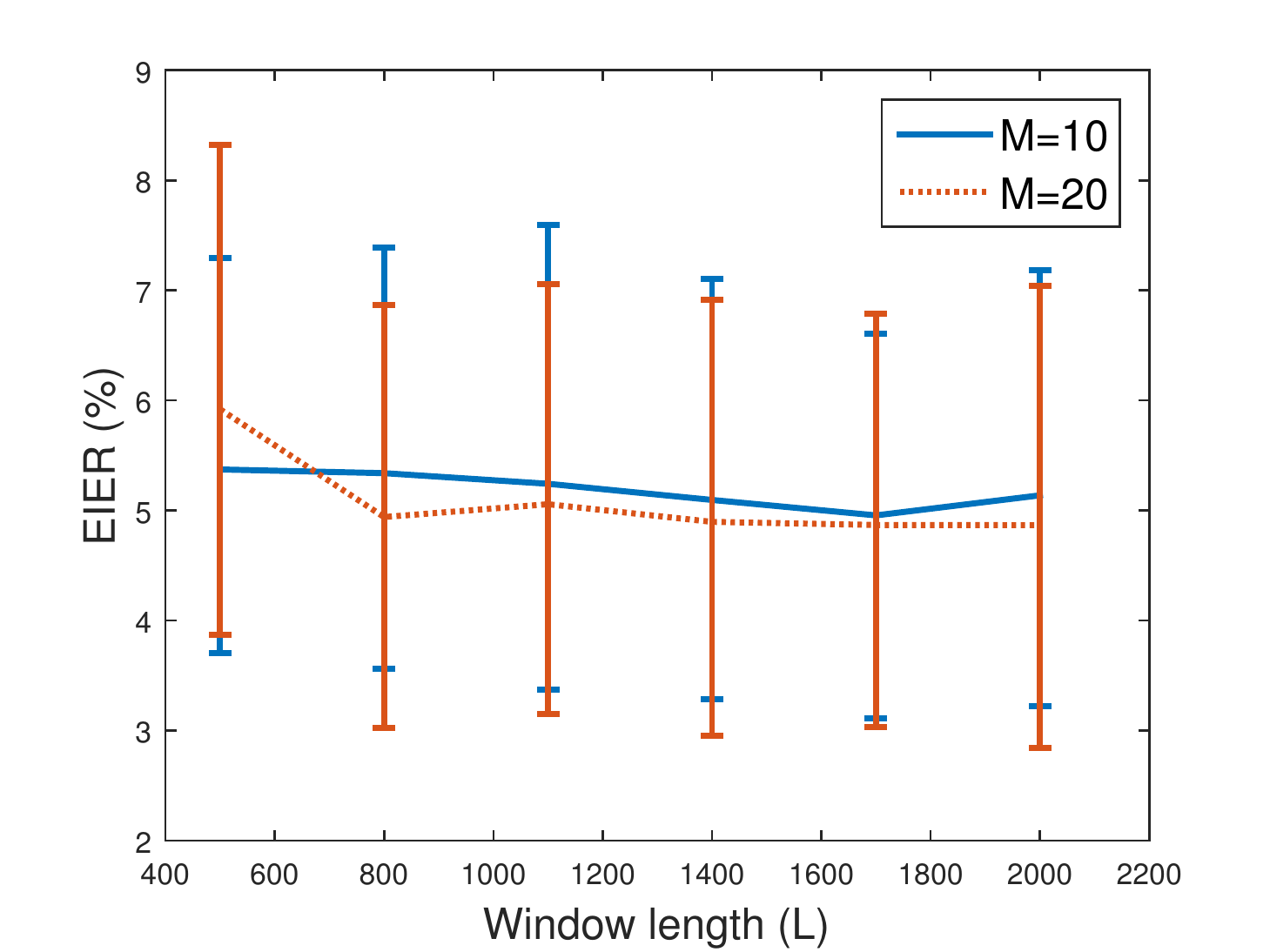}
\centerline{(b)}
\end{minipage}
 \caption{EIER for different window lengths, with: a) $\Omega=\{(i,j)| i=1,\ldots, N, j=1,\ldots, M\}$; b) $50
\%$ misses in $\bbR^x$; c) $\Omega=\emptyset$.} 
 \label{fig:EIER}
\end{figure*}

%
%
%
%

 \noindent
\textbf{Results.} Figure~\ref{fig:heatmap} depicts actual and inferred adjacency matrices, resulting from one realization of Algorithm~\ref{algo:netid} for $M \in \{10,20 \}$, with $L = 1,000$ per experiment. As shown in the plot, fewer edges are erroneously identified as the number of windows $M$ increases. This is not really surprising because the probability that the condition in Theorem \ref{theorem:full} is satisfied will improve with larger $M$. Figures~\ref{fig:EIER} plots EIER values against $L$, averaged over $500$ independent runs of Algorithm~\ref{algo:netid} for $M=10$ and $M=20$. 

Figure~\ref{fig:EIER}(a) plots the observed error performance over several window lengths ($L$), when $\bbR^x$ is fully available, whereas Figure~\ref{fig:EIER}(b) was obtained after random omission of entries in $\bbR^x$ with probability $0.5$. On the other hand, Figure~\ref{fig:EIER}(c) depicts performance in the completely blind case, that is, $\Omega=\emptyset$. In all three scenarios, there is a general increase in edge identification accuracy with $L$, since wider window lengths yield improved estimates of the correlation matrices per window. Not surprisingly, the semi-blind topology inference approach in Section \ref{sec:identifiability}-B outperforms the completely blind alternative $(\Omega=\emptyset)$, since one presumably has more prior information available. On the other hand, in the completely blind case, Algorithm \ref{algo:netid} still results in a reliable estimate of the network topology with low edge identification error.
 
\begin{figure}[!t]
 \centering
\begin{tabular}{cc}
\hspace*{-4ex}
\includegraphics[width=4.7cm]{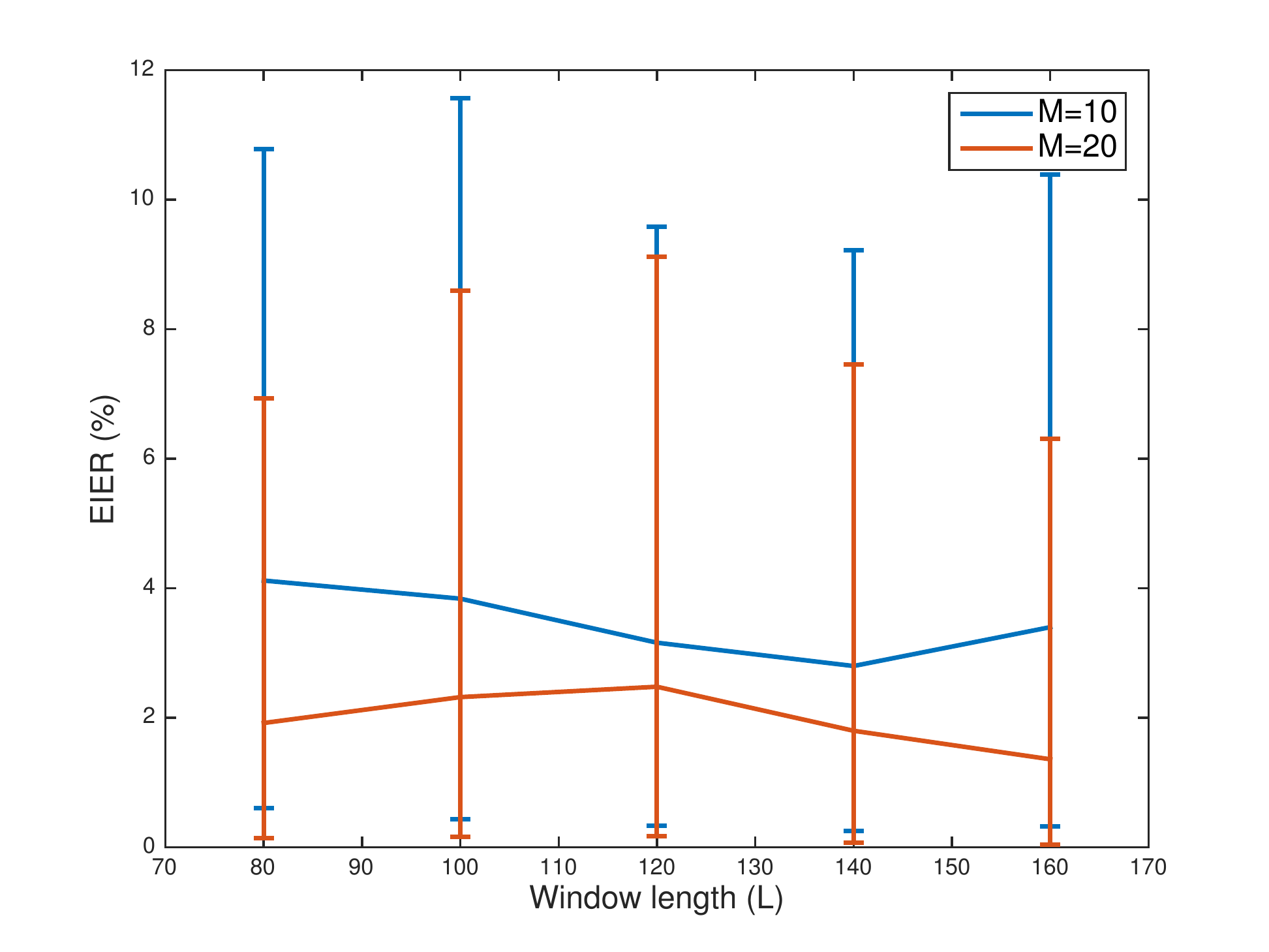}&
\hspace*{-5ex}
\includegraphics[width=4.7cm]{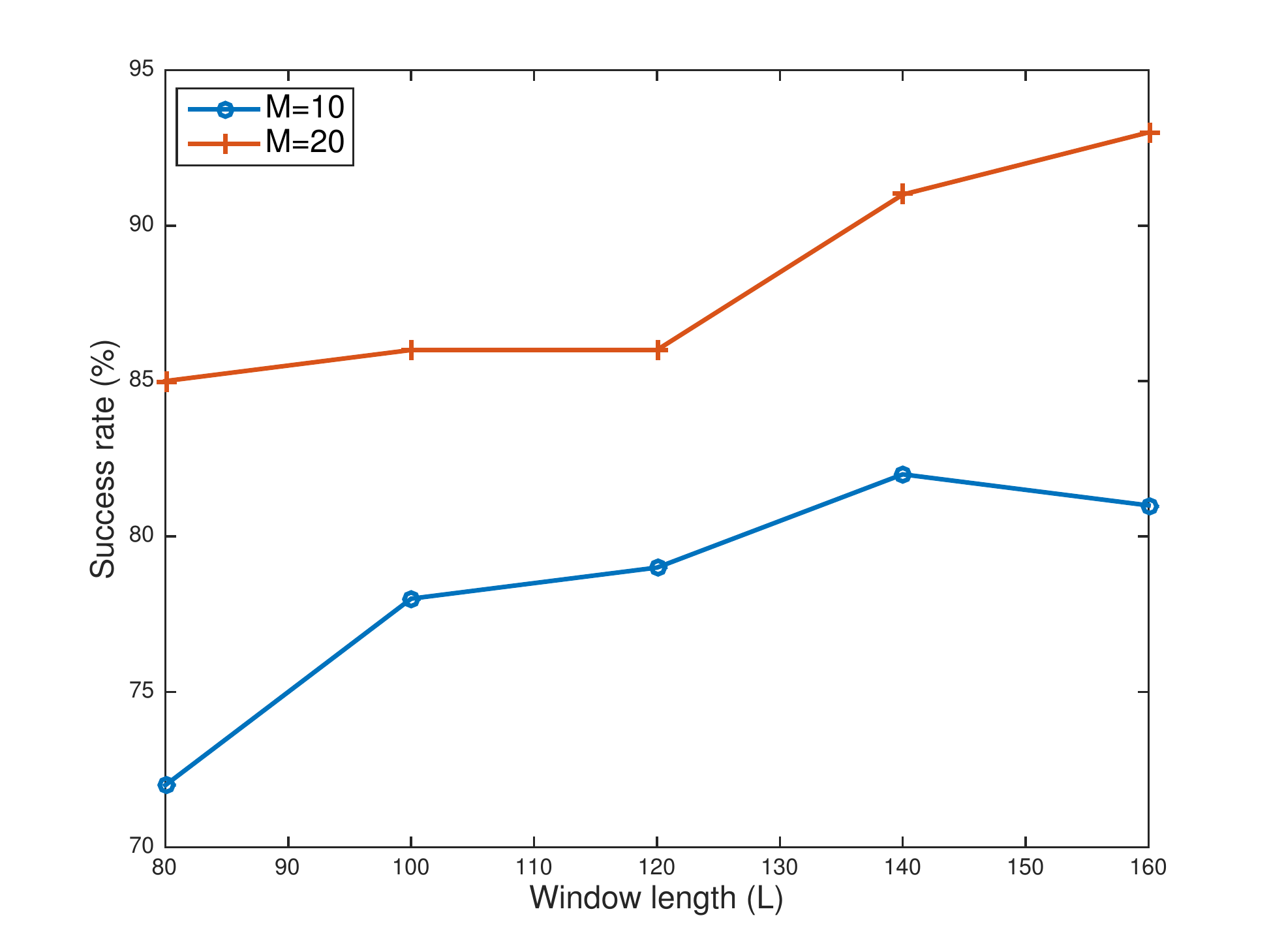}
\\
(a)& (b)
\end{tabular}
  \caption{Performance in blind scenario: a) EIER; b) Success rate.}
\label{fig:renyi}
\end{figure}

In several real-world applications, exogenous variables are often unavailable or costly to measure, hence performance benchmarks for the developed algorithm in such blind settings are of considerable interest. To facilitate further assessment of the stability of the novel algorithm when operating in blind scenarios, an extended experiment was carried out as follows. Per experiment trial, an unweighted Erd\"{o}s-Renyi random graph with $5$ nodes was generated, with the probability that any node pair is connected by an edge set to $0.4$, and then Algorithm~\ref{algo:netid} was run with $\Omega=\emptyset$. For this experiment, Figure~\ref{fig:renyi} (a) depicts the resulting EIER performance, averaged over $100$ independent runs. Figure~\ref{fig:renyi} (b) depicts  the success rate of the experiments, with a trial is considered successful if $ \text{EIER} =  0$. It is clear from the results that the majority of trials succeeded in exact identification of all edges. This is an exciting empirical result that demonstrates the potential for the proposed algorithm to provide reliable estimates in blind settings, even under the presence of noise. The implications of this empirical result  are well-motivated in real-world applications, where exogenous inputs are unavailable to eliminate the inherent permutation ambiguity.

\subsection{Simulated piecewise-constant network}
\label{subsec:sim_pconst}
\noindent\textbf{Data generation.} An initial $64$-node network was generated with adjacency matrix $\bbA_0$ via the Kronecker random graph model, as detailed in the previous subsection. Edge weights in the initial non-zero support of $\bbA_0$ were varied over time windows, following two edge-variation patterns: p1) $a_{ij}^m=a_{ij}^0+0.1\text{sin}(0.01m)$, for $m=1,\dots,200$; and p2) $a_{ij}^m=0$ with probability 0.2 at the $50$th and $100$th time windows. For $L=500$, $L=2,000$, and $L=3,000$, endogenous measurements were simulated over $T=ML$ time-slots,  partitioned into $M$ windows of fixed length $L$. The window boundaries were preselected as $\{ \tau_m \}_{m=1}^{M+1}$, with $\tau_1 = 1$ and $L := \tau_{m+1} - \tau_{m}$. Per $t \in [\tau_{m}, \tau_{m+1} - 1]$, exogenous inputs were sampled as $\bbx_t \sim \mathcal{N}(\boldsymbol{0}, \sigma_m^2 \bbI)$, with $\{ \sigma_m^2 \}_{m=1}^M$ set to $M$ distinct values. With $\mathbf{e}_t$ sampled i.i.d. from $\mathcal{N}(\mathbf{0}, 10^{-2} \bbI)$, $\bby_t$ was similarly generated using the SEM, that is, $\bby_t = (\bbI - \bbA_m)^{-1}(\mathbf{B}\bbx_t + \mathbf{e}_t)$, where $[\bbB]_{jj} \sim \mathcal{U}[2, 3]$. 

\begin{figure}[!t]
 \centering
\begin{tabular}{cc}
\hspace*{-4ex}
\includegraphics[width=4.7cm]{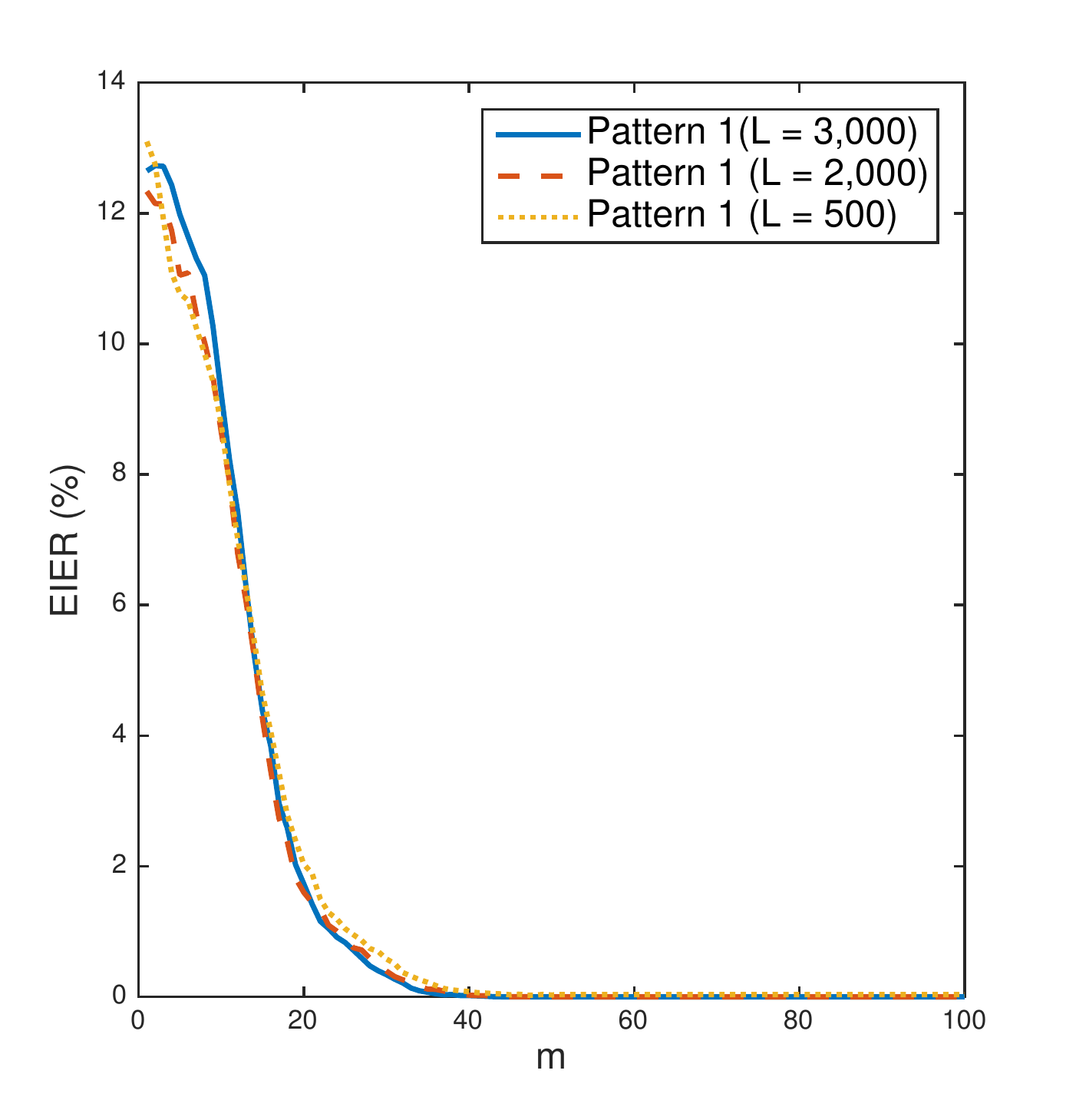}&
\hspace*{-5ex}
\includegraphics[width=4.7cm]{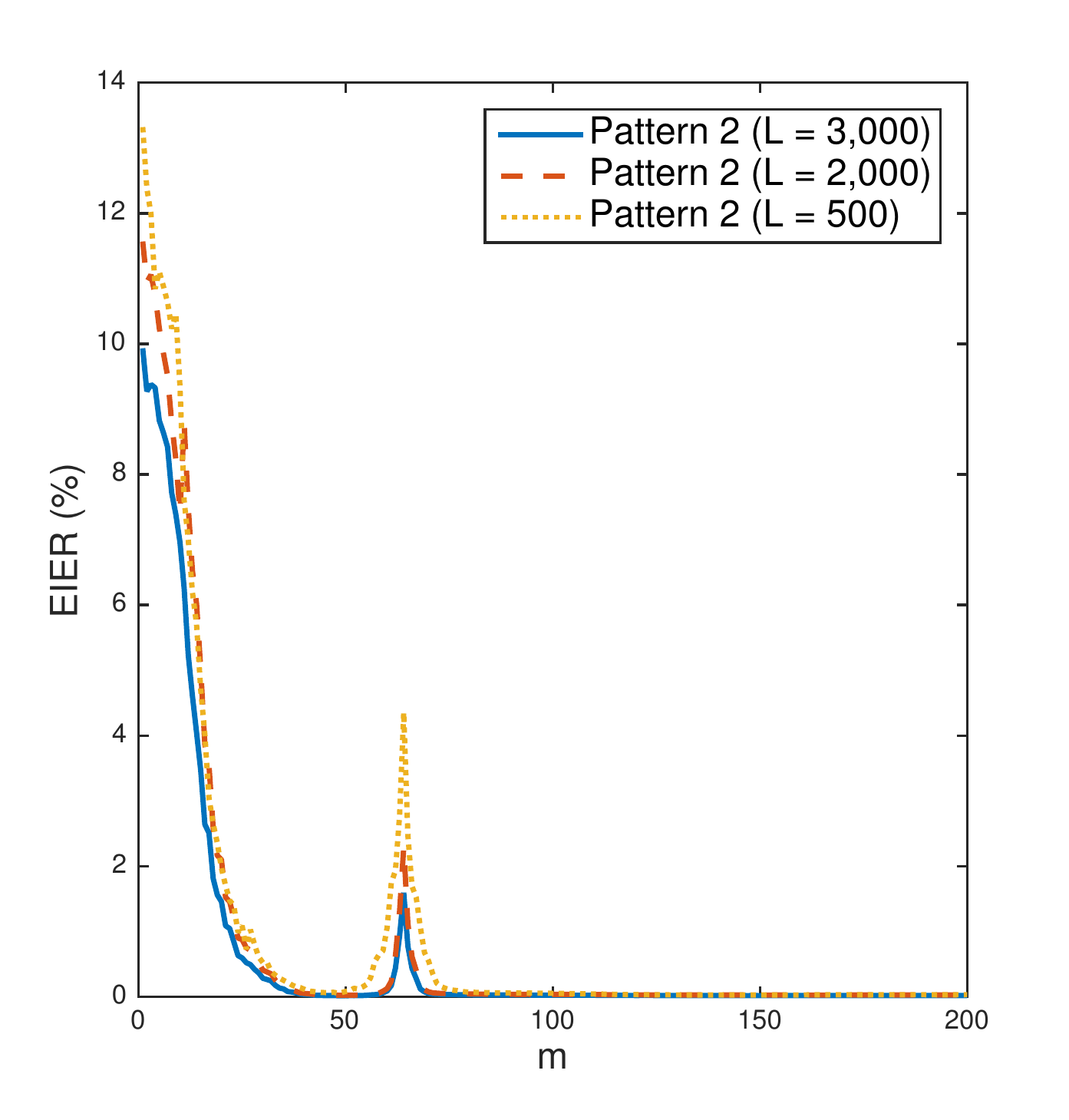}
\\
(a)& (b)
\end{tabular}
  \caption{EIER vs. $m$ for: (a) Scenario p1; and (b) Scenario p2.}
\label{fig:EIER:dynamic}
\end{figure}

\begin{figure}[!t]
 \centering
\begin{tabular}{cc}
\hspace*{-4ex}
\includegraphics[width=4.7cm]{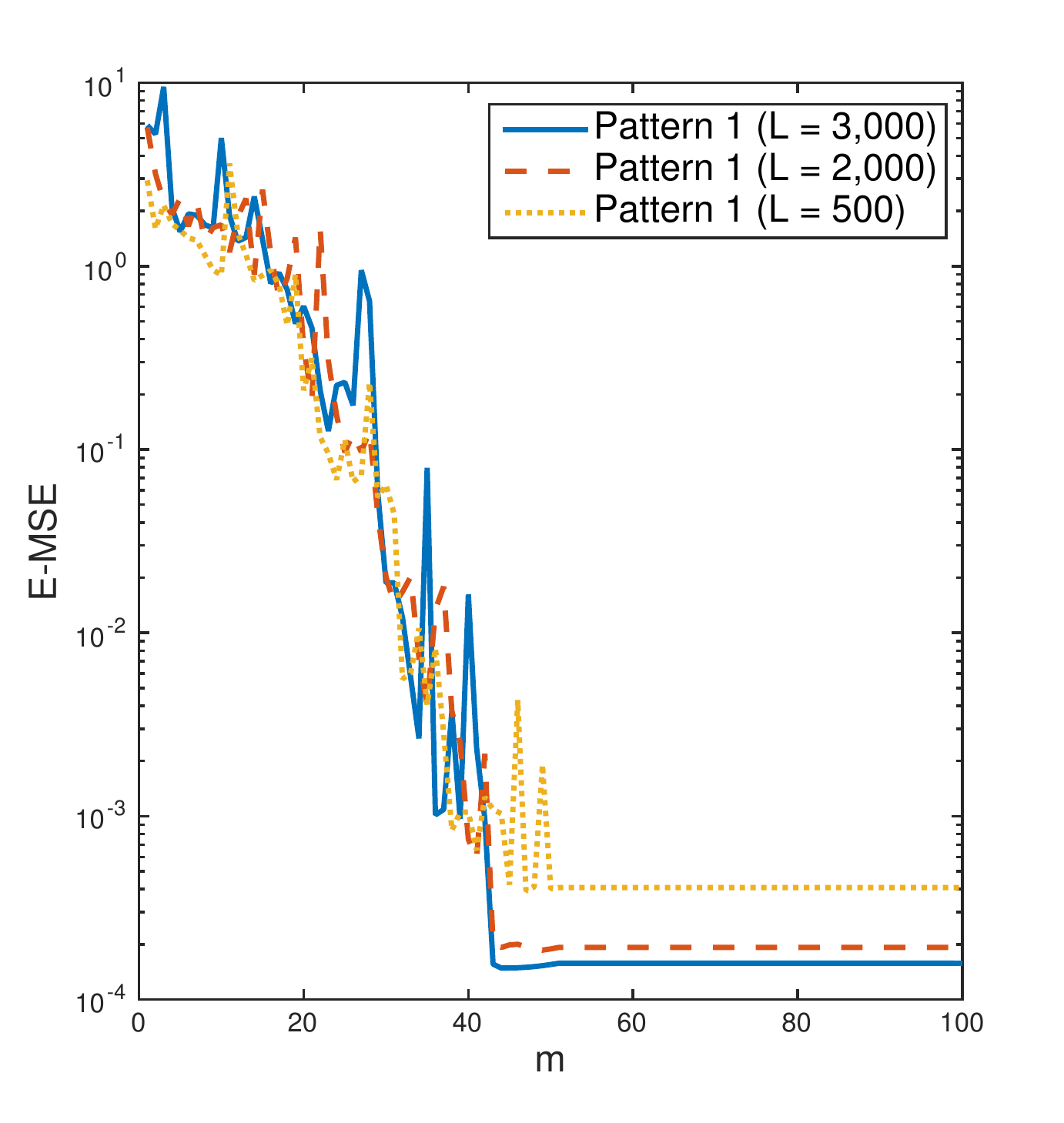}&
\hspace*{-5ex}
\includegraphics[width=4.7cm]{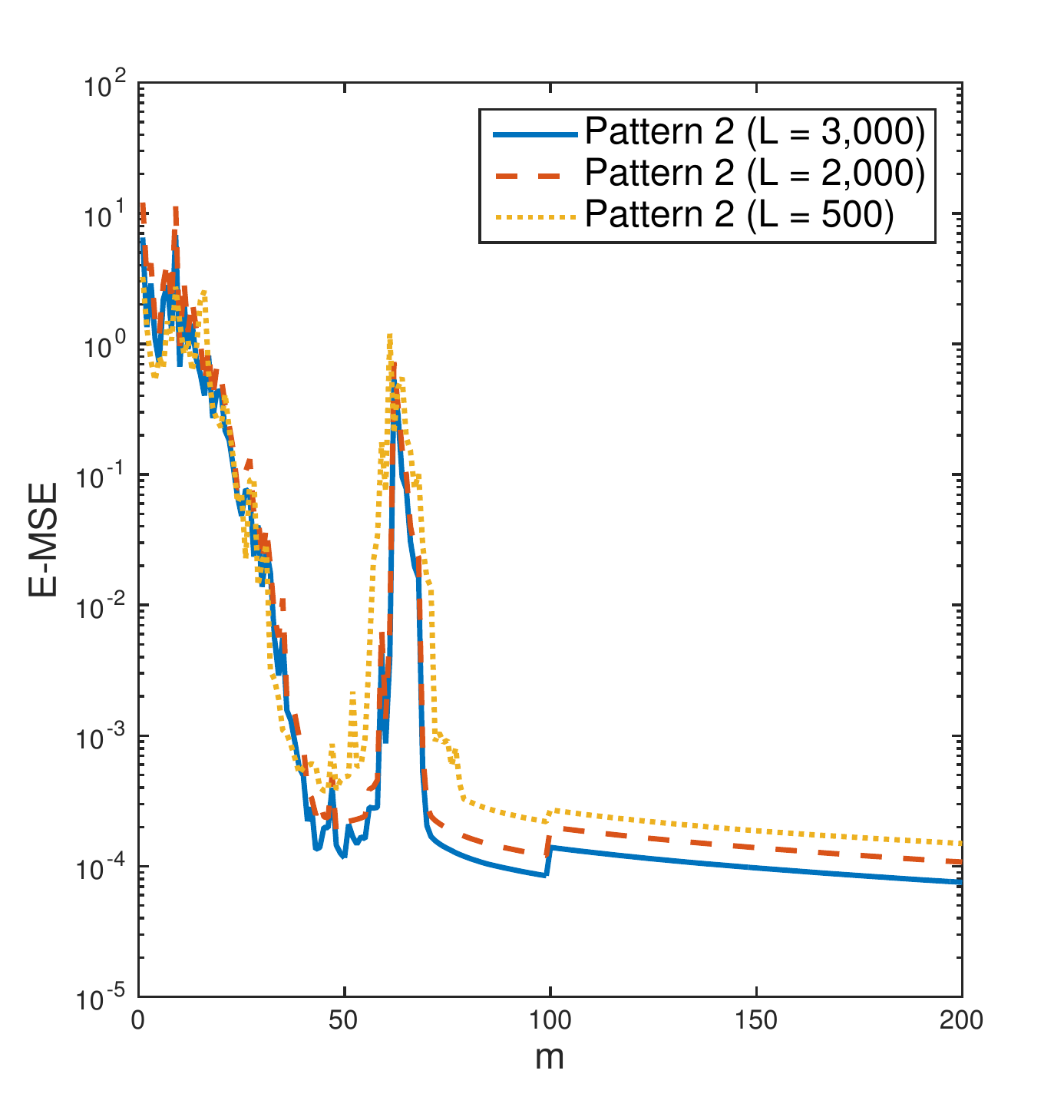}
\\
(a)& (b)
\end{tabular}
  \caption{MSE vs $m$ for: a) Scenario p1; b) Scenario p2.}
\label{fig:NMSE:dynamic}
\end{figure}

\begin{figure}[h]
	\centering
	\hspace*{-4ex}
\includegraphics[width=10cm]{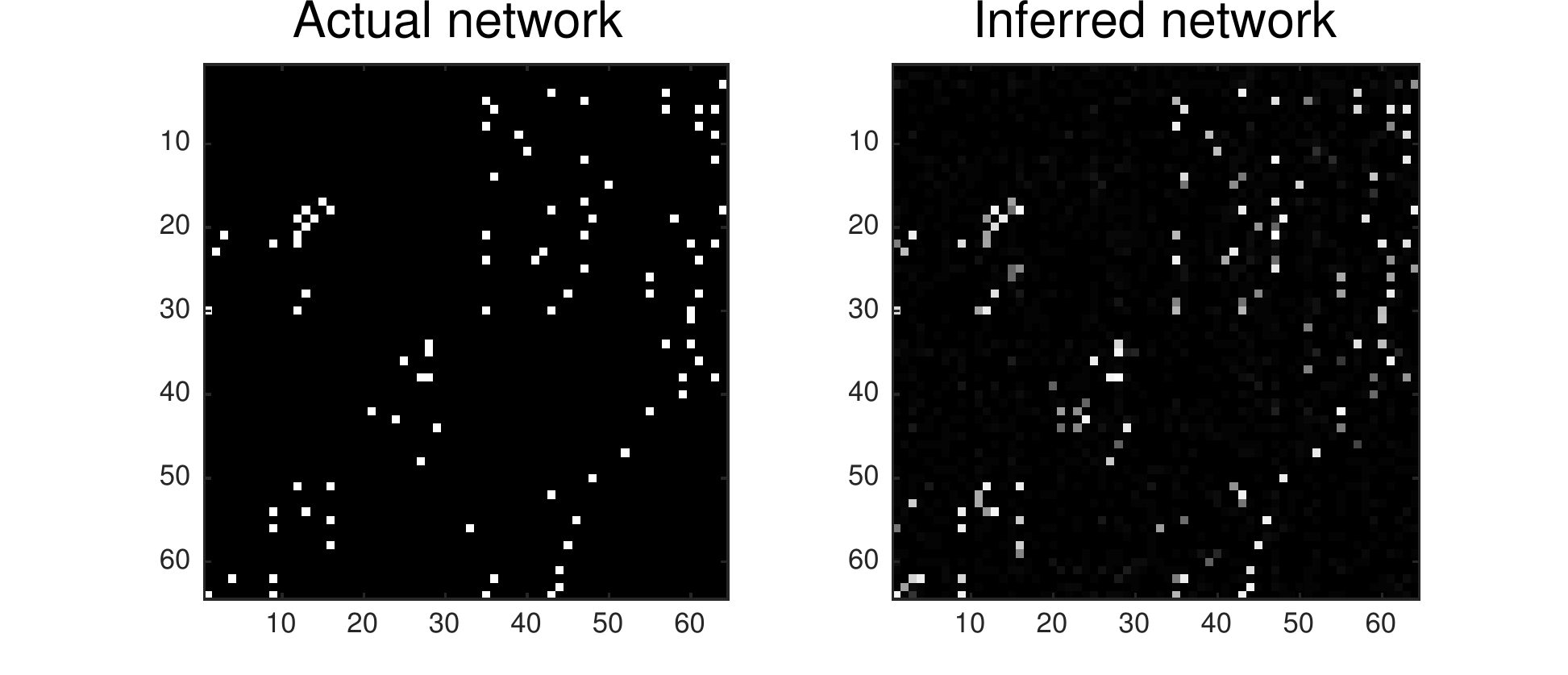}
\caption{Actual and inferred networks at $m=200$.}
\label{fig:heatmap2}
\end{figure}

 \noindent\textbf{Results.} Algorithm~\ref{algo:netid:online} was run on the simulated data using $\beta = 0.999$, with an edge declared present if $\hat{a}_{ij}$ exceeded a threshold $\eta$ set to yield the lowest EIER. Algorithm performance was assessed with respect to both EIER, and the empirical mean-square error (E-MSE), defined as $\text{E-MSE}:=\|\bbA_m-\hat{\bbA}_m\|_F^2/(N(N-1))$. In addition, both error metrics were averaged over $100$ runs per experiment. 

As shown by both Figures~\ref{fig:EIER:dynamic} and \ref{fig:NMSE:dynamic}, Algorithm~\ref{algo:netid:online} tracks the evolution of the network remarkably well. During windows where the edge support is known to change, error metrics increase in value, but gracefully return to lower values. Figure~\ref{fig:heatmap2} depicts heatmaps of actual and inferred adjacency matrices, obtained by running Algorithm~\ref{algo:netid:online} during the window indexed by $m=200$ for scenario p2).

\begin{figure*}[tpb!]
\begin{minipage}[b]{.48\textwidth}
\centering
\includegraphics[width=9cm]{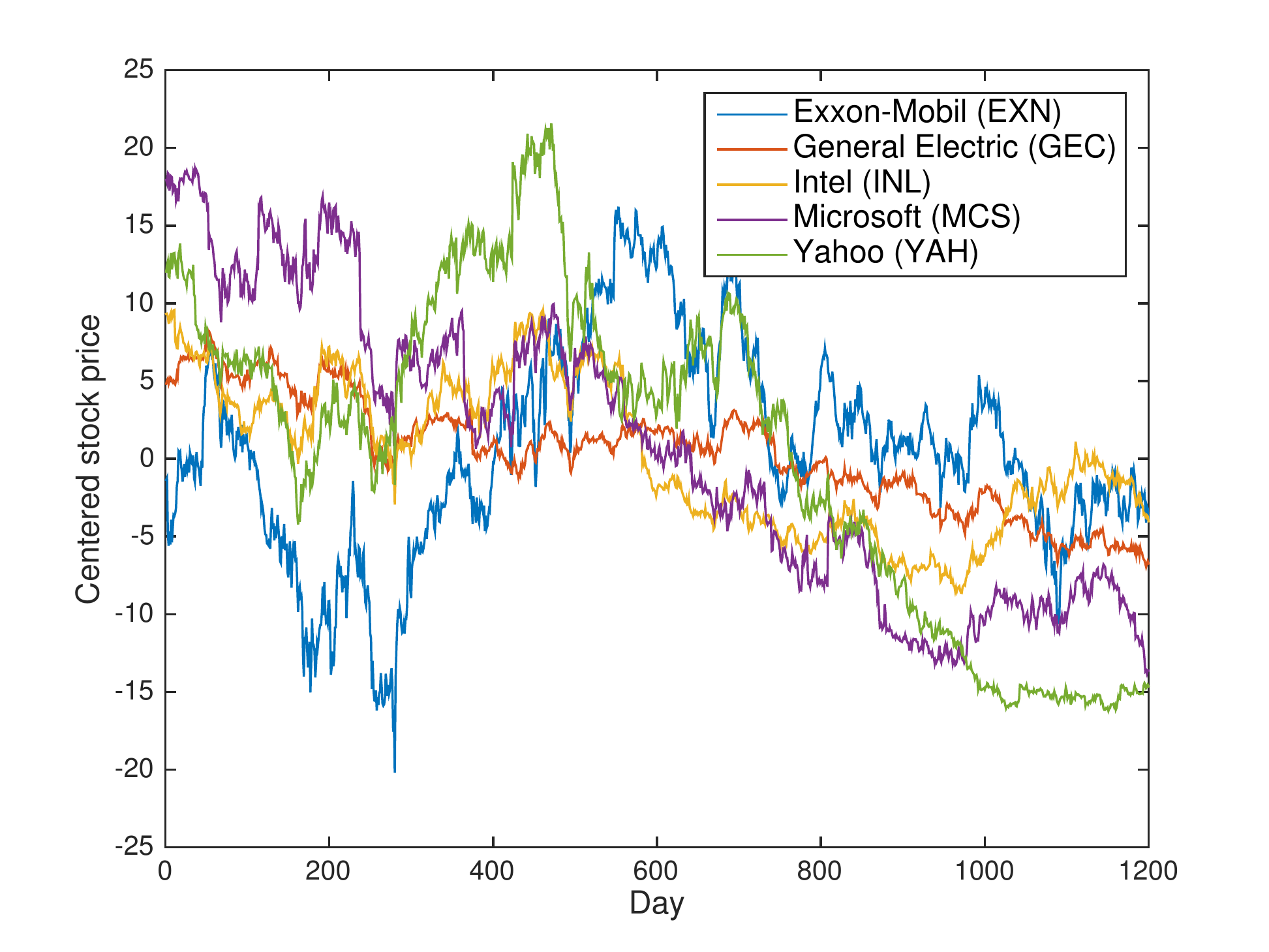}
\centerline{(a)}
\end{minipage}
\begin{minipage}[b]{.48\textwidth}
\centering
\includegraphics[width=9cm]{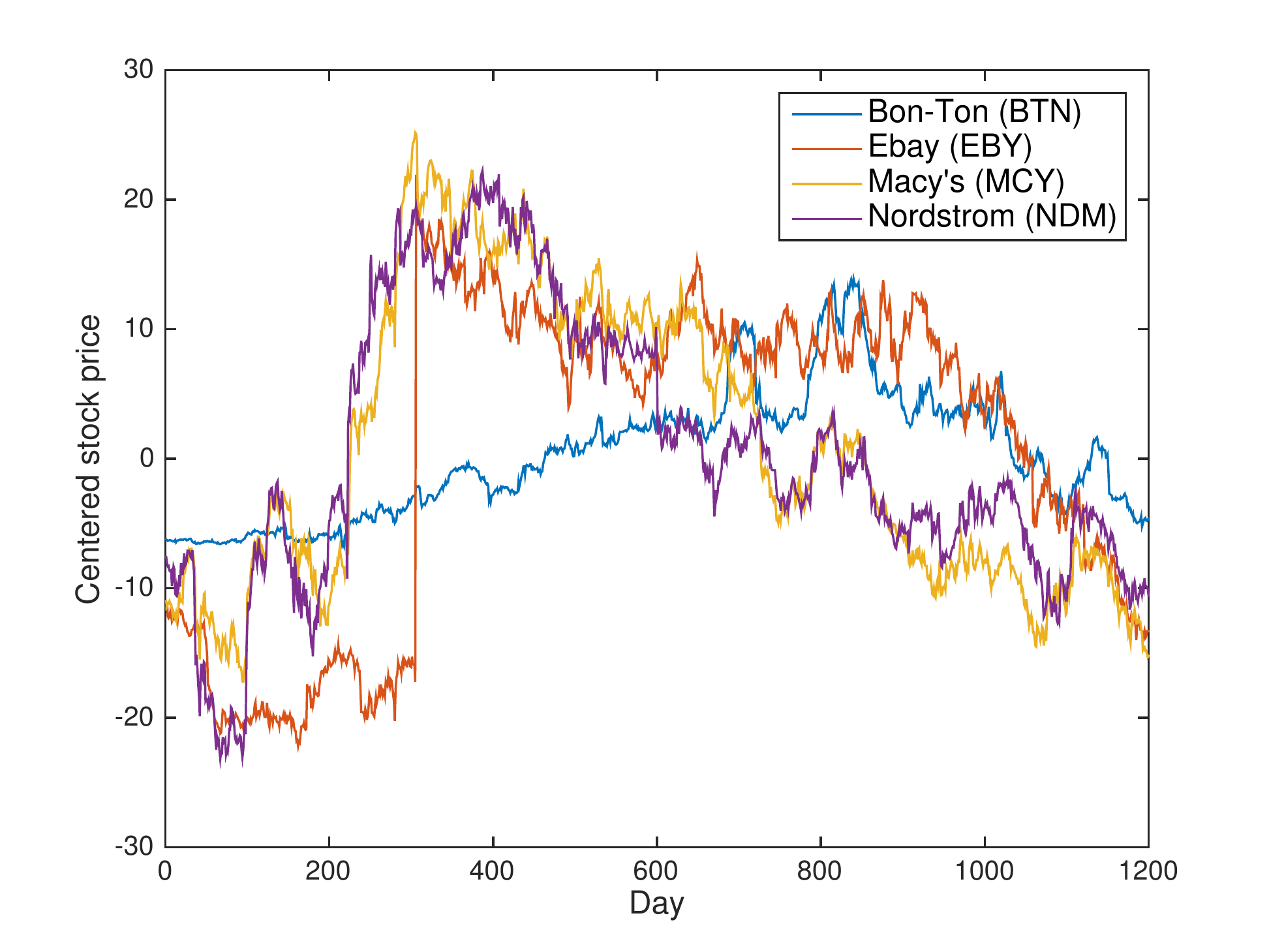}
\centerline{(b)}
\end{minipage}
 \caption{Plot of the two groups of stock prices over the observation duration with zero-mean centering: a) technology companies; and b) online and ``brick-and-mortar'' retailers. The stock ticker symbol for each company is shown in the legend (in parentheses). 
 } 
 \label{fig:price}
\end{figure*}

\begin{figure*}[tpb!]
\begin{minipage}[b]{.48\textwidth}
\centering
\includegraphics[width=8cm]{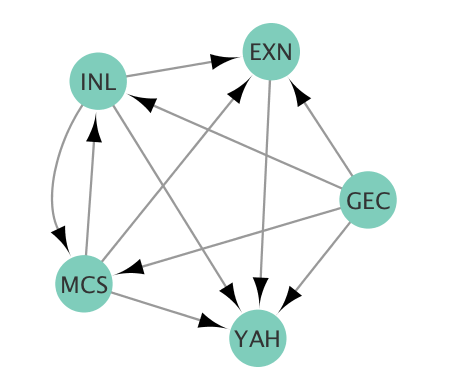}
\centerline{(a)}
\end{minipage}
\begin{minipage}[b]{.48\textwidth}
\centering
\includegraphics[width=8cm]{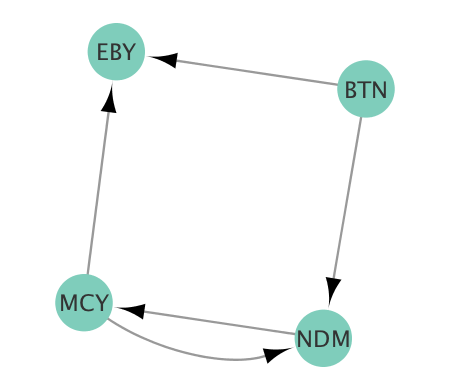}
\centerline{(b)}
\end{minipage}
 \caption{Visualization of network topologies inferred from the stock price time series, depicting: a) technology companies; and b) online and ``brick-and-mortar'' retailers. Notice the stronger dependencies between the two competing ``brick-and-mortar'' retailers, Macy's (MCY) and Nordstrom (NDM).} 
 \label{fig:stocknet}
\end{figure*}

\subsection{Tests on real networks}

\noindent\textbf{Data description.} To conduct tests on real-world networks, historical stock price data were downloaded through a free \emph{Yahoo} application program interface (API). Historical closing prices were obtained as time series for dates ranging from December $23$, $2011$ to September $30$, $2016$ ($1,200$ days in total). The stock time series were grouped into two clusters, namely: a) large technology companies (\emph{Exxon-Mobil, Intel, Microsoft, Yahoo, and General Electric}), and b) online and brick-and-mortar retailers (\emph{Bon-Ton, E-bay, Macy's, and Nordstrom}). Choices of which stocks were classified under the two groups were based on prior knowledge of historical inter-dependencies existing among them in financial markets. For instance, a significant drop in Intel stock prices often signals changes in share prices for Microsoft, Intel, and sometimes General Electric.

 \noindent\textbf{Results.} For this set of experiments, the combined multivariate time series were adopted as endogenous variables $\left( \{\bby_t\}_{t=1}^{1,200} \right) $, after a pre-processing step in which they were centered to have zero mean; see Figure \ref{fig:price} for a plot of the centered time series. Furthermore, money invested in the stocks constitutes exogenous inputs $\left( \{\bbx_t\}_{t=1}^{1,200} \right)$, which are not known in this case, since such information is generally not privy to the public, hence $\Omega=\emptyset$. Furthermore, it was observed that most stock prices tend to exhibit steady quarterly trends (rising or falling), and the window length was consequently set to $L=100$ for all tests. Algorithm \ref{algo:netid} was then run with $\Omega=\emptyset$, and $M=12$ to infer the causal dependencies between the selected stock prices.

According to the discussion in Section~\ref{sec:identifiability}, there is no guarantee of identifiability in the completely blind setting. Fortunately, the simulated tests depicted by Figure~\ref{fig:renyi} demonstrate that when the network has a few nodes, there is a high probability of successful recovery of the true network in the presence of noise. Based on this empirical observation, it is reasonable to expect that if only a few stocks are selected, then many trials will yield the true network upon running Algorithm~\ref{algo:netid} with random initializations. To this end, $100$ independent runs of Algorithm~$\ref{algo:netid}$ were done with random initializations, and it turned out that most estimates yielded the same support for $\hat{\mathbf{A}}$, with very slight variations in actual values of its entries. Consequently, a simple scheme was adopted to infer the network topology from the ensemble of estimates. Unique topologies based on the support of  $\hat{\mathbf{A}}$ for the $100$ realizations were enumerated, and a majority voting scheme was adopted to reach consensus on the final topology. The most frequent network topologies from the experiments are depicted by Figure~\ref{fig:stocknet}, with (a) representing a majority vote of  $92$ out of $100$, while (b) was the result inferred from $68$ experiments. The figure shows very strong dependencies in the first group of technology companies, while the second plot shows stronger inter-dependencies between Macy's and Nordstrom than the others. Interestingly, both Macy's and Nordstrom are well-known ``brick-and-mortar'' retailers and competitors. The stronger dependence between them seems to agree with the expectation that changes in the price of one would be expected to indirectly impact the other.

%
%

\section{Conclusions}
\label{sec:conc}

This paper put forth a novel approach for inference of network topologies from the statistics of nodal processes. Leveraging SEMs, the network topology inference task was reformulated as a constrained PARAFAC tensor decomposition. Recognizing the inherent uniqueness challenges, conditions under which the network can be uniquely identified were derived. Unlike conventional SEMs, which require exact information of the exogenous inputs in order to guarantee identifiability, it was proven that the novel tensor-based approach is capable of uniquely identifying the network topology only with partial information of the second-order statistics of nodal exogenous inputs.

The framework was further extended to facilitate real-time sequential estimation of the network topology by developing a novel topology tracking algorithm. An exponentially weighted least-squares estimator was advocated for the topology tracking problem, making it possible to efficiently solve the problem ``on the fly.'' To assess the effectiveness of the novel approaches, extensive numerical tests were conducted on both simulated data and historical stock prices of several publicly-traded corporations. 

In order to broaden the scope of this work, there are several intriguing directions for future investigation, namely: a) developing algorithms that are capable of exploiting prior knowledge pertaining to the network structure e.g., edge sparsity or power law degree distributions; and b) distributed implementation of the novel algorithms, which is well-motivated, especially when dealing with large-scale networks.

\section*{Appendix}
\renewcommand{\thesubsection}{\Alph{subsection}}

\subsection{Proof of Proposition \ref{proposition1}} 
\label{appendix2}
Since diagonal entries of $\bbA$ are all zero, and $\bbB^{-1}$ is a diagonal matrix with nonzero entries, $\boldsymbol{\mathcal{A}}$ is invertible; that is,
\begin{align}
\label{eq:app_a0}
	\boldsymbol{\mathcal{A}}^{-1}=\bbB^{-1}(\bbI-\bbA).
\end{align}
Clearly, the diagonal entries of $\boldsymbol{\mathcal{A}}^{-1}$ coincide with those of $\bbB^{-1}$, which implies that
\begin{align}
\label{app:1}
	\bbB =\left(\text{Diag}\big[\boldsymbol{\mathcal{A}}^{-1}\big]\right)^{-1}.
\end{align}
Recognizing that $\bbB\boldsymbol{\mathcal{A}}^{-1}=\bbI-\bbA$, one can write 
\begin{align}
	\bbA = \bbI-\bbB\boldsymbol{\mathcal{A}}^{-1}=\bbI-\left(\text{Diag}(\boldsymbol{\mathcal{A}}^{-1})\right)^{-1}\boldsymbol{\mathcal{A}}^{-1}
\end{align}
which completes the proof.

%
%
%
%

\subsection{Proof of Lemma \ref{lemma1}}
\label{app:lemmac}
First, note that \eqref{eq:lemmac1} can be written as
	\begin{align}
	\label{eq:id:1}
		\bbR^x-\bbR^x\bbPi\bbLambda_3=\boldsymbol{0}_{M \times N}
	\end{align}
and recall that $\bbPi$ is a permutation matrix; hence, each constituent column in $\bbPi$ comprises zeros with the exception of a single entry set to one. 
Letting $\pi_{ij}$ denote the $(i, j)$-th entry of $\bbPi$, assume without loss of generality that $\pi_{ij}=1$ and $\pi_{kj}=0, ~\forall k\neq i$. Consequently, with $\bbp_j\in\mathbb{R}^{N}$ representing column $j$ of $\bbP:=\bbPi\bbLambda_3$, one can equivalently write
\begin{align}
\label{def:p}
	\bbp_j=[0,\ldots, 0,\underbrace{ \pi_{ij}\lambda_{j}}_{\text{entry } i}, 0, \ldots,0]^\top
\end{align}
where $\lambda_j\neq 0$ denotes the $j$-th diagonal entry of $\bbLambda_3$.
 Extracting the $j$-th column on both sides of \eqref{eq:id:1}, namely,
\begin{align}
\label{eq:id2}
	\bbr^x_j-\bbR^x\bbp_j=\boldsymbol{0}_{M\times 1}
\end{align}
and combining \eqref{def:p} and \eqref{eq:id2}, one obtains
\begin{align}
\label{eq:id3}
	\bbr^x_j=\pi_{ij}\lambda_j\bbr^x_i.
\end{align}
When $i\neq j$, \eqref{eq:id3} implies that $\bbr^x_i$ and $\bbr^x_j$ are linearly dependent, which contradicts the condition $\text{kr}({\bbR^x})\geq 2$ in Lemma \ref{lemma1}. Hence, for \eqref{eq:id3} to hold for some nonzero $\lambda_j$, it is necessary that $i=j$, which is equivalent to requiring $\pi_{jj}=1$ and $\lambda_j=1$. Since this holds for any $j$, one deduces that
\begin{align}
	\bbPi=\bbI, ~~
	\bbLambda_3=\bbI.
\end{align}
\subsection{Proof of Lemma \ref{lemma2}}
\label{app:lemma2}
Recalling from Algorithm~\ref{algo:netid} that
\begin{eqnarray}
	\hat{\bbA}
	&=&\bbI-\left(\text{Diag}(\widehat{\boldsymbol{\mathcal{A}}}^{-1})\right)^{-1}\widehat{\boldsymbol{\mathcal{A}}}^{-1}\nonumber
\end{eqnarray}
and substituting $\widehat{\boldsymbol{\mathcal{A}}}=\boldsymbol{\mathcal{A}} \bbLambda$, one obtains
\begin{eqnarray}	
\hat{\bbA} 
	&=&\bbI-\left(\text{Diag}\big[(\boldsymbol{\mathcal{A}}\bbLambda)^{-1}\big]\right)^{-1}(\boldsymbol{\mathcal{A}}\bbLambda)^{-1}\nonumber\\
	& = &\bbI-\left(\text{Diag}\big[(\boldsymbol{\mathcal{A}})^{-1}\big]\right)^{-1}\bbLambda\bbLambda^{-1}\boldsymbol{\mathcal{A}}^{-1}\nonumber\\
	& = & \bbI-\left(\text{Diag}(\boldsymbol{\mathcal{A}}^{-1})\right)^{-1}\boldsymbol{\mathcal{A}}^{-1}.
\end{eqnarray}
Comparing with Proposition \ref{proposition1}, it is clear that $ \hat{\bbA}=\bbA\nonumber $, which concludes the proof.

\subsection{Proof of Lemma \ref{lemma3}}
\label{app:lemmac:s}
First, assume without loss of generality that column $j$ of the permutation matrix $\check{\bbPi}$ satisfies $\check{\pi}_{ij}=1$ and $\check{\pi}_{kj}=0, ~\forall k\neq i$, with $\check{\pi}_{ij}$ denoting entry $(i, j)$ of $\check{\bbPi}$. Since $\check{\bbp}_j\in\mathbb{R}^{N}$, the $j$-th column  of $\check{\bbP}:=\check{\bbPi}\check{\bbLambda}_3$ can be written as
\begin{align}
\label{def:p:s}
	\check{\bbp}_j:=[0,\ldots 0,\underbrace{ \check{\pi}_{ij}\check{\lambda}_{j}}_{\text{entry } i},0,\ldots,0]^\top
\end{align}
with $\lambda_j\neq 0$ representing the $j$-th diagonal entry of $\bbLambda_3$.
Extracting entries indexed by $\Omega_i\cup \Omega_j$ in column $j$ on both sides of \eqref{eq:lemmac1s}, one has
\begin{align}
\label{eq:id3s}
	\check{\bbr}_{j}^i=\check{\pi}_{ij}\check{\lambda}_j\check{\bbr}_{i}^j
\end{align}
and assuming that $i\neq j$, \eqref{eq:id3s} implies that $\check{\bbr}_{i}^j$ and $\check{\bbr}_{j}^i$ are linearly dependent, which contradicts the condition in Theorem~\ref{theorem:full}. As a result, for \eqref{eq:id3s} to hold true for some nonzero $\lambda_j$, it is necessary that $i=j$, which is equivalent to having $\check{\pi}_{jj}=1$ and $\check{\lambda}_j=1$. Recognizing that this holds for any $j$, one arrives at
\begin{align}
	\check{\bbPi}=\bbI, ~~
	\check{\bbLambda}_3=\bbI.
\end{align}

\section*{Acknowledgment}
The work of this paper was initiated in a project of a course taught by Prof. N. D. Sidiropoulos, and the authors would like to thank him for his feedback.
\balance
%
\bibliography{net,myabrv,tensor}
\bibliographystyle{IEEEtranS}
\end{document}